\documentclass[10pt,twocolumn,letterpaper]{article}

\usepackage[pagenumbers]{cvpr} 

\usepackage{graphicx}
\usepackage{amsmath}
\usepackage{amssymb}
\usepackage{booktabs}

\makeatletter
\@namedef{ver@everyshi.sty}{}
\makeatother

\usepackage[pagebackref,breaklinks,colorlinks]{hyperref}

\usepackage[capitalize]{cleveref}
\crefname{section}{Sec.}{Secs.}
\Crefname{section}{Section}{Sections}
\Crefname{table}{Table}{Tables}
\crefname{table}{Tab.}{Tabs.}

\def\cvprPaperID{*****} 
\def\confName{CVPR}
\def\confYear{2022}

\usepackage{overpic}
\usepackage{enumitem} 
\usepackage{overpic} 
\usepackage{color}

\usepackage{everyshi}
\usepackage{adjustbox}
\usepackage{multirow}
\usepackage[dvipsnames, table]{xcolor}
\definecolor{anti-flashwhite}{rgb}{0.95, 0.95, 0.96}
\definecolor{whitesmoke}{rgb}{0.94, 0.94, 0.94}
\definecolor{teagreen}{rgb}{0.82, 0.94, 0.75}
\definecolor{powderblue}{rgb}{0.69, 0.88, 0.9}
\definecolor{pastelblue}{rgb}{0.68, 0.78, 0.81}
\definecolor{lightskyblue}{rgb}{0.53, 0.81, 0.98}

\usepackage[absolute,overlay]{textpos}
\definecolor{turquoise}{cmyk}{0.65,0,0.1,0.3}
\definecolor{purple}{rgb}{0.65,0,0.65}
\definecolor{dark_green}{rgb}{0, 0.5, 0}
\definecolor{orange}{rgb}{0.8, 0.6, 0.2}
\definecolor{red}{rgb}{0.8, 0.2, 0.2}
\definecolor{darkred}{rgb}{0.6, 0.1, 0.05}
\definecolor{blueish}{rgb}{0.0, 0.3, .6}
\definecolor{light_gray}{rgb}{0.7, 0.7, .7}
\definecolor{pink}{rgb}{1, 0, 1}
\definecolor{greyblue}{rgb}{0.25, 0.25, 1}

\newcommand{\Fig}[1]{Fig.~\ref{fig:#1}}

\newcommand{\Tab}[1]{Tab.~\ref{tab:#1}}
\newcommand{\Table}[1]{Table~\ref{tab:#1}}

\usepackage{blindtext}

\renewcommand{\paragraph}[1]{\vspace{1em}\noindent\textbf{#1}.}
\begin{document}

\title{MPViT : Multi-Path Vision Transformer for Dense Prediction}

\newcommand*{\affaddr}[1]{#1} 
\newcommand*{\affmark}[1][*]{\textsuperscript{#1}}
\newcommand*{\email}[1]{\texttt{#1}}

\author{%
Youngwan Lee\affmark[1,2]~~~Jonghee Kim\affmark[1]~~~Jeff Willette\affmark[2]~~~Sung Ju Hwang\affmark[2,3]\\ 
\vspace{-0.1in}
\\
\affaddr{\affmark[1]Electronics and Telecommunications Research Institute~(ETRI), South Korea}\\
\affaddr{\affmark[2]Korea Advanced Institute of Science and Technology~(KAIST), South Korea}\\
\affaddr{\affmark[3]AITRICS, South Korea}\\
}



\maketitle


\begin{abstract}
Dense computer vision tasks such as object detection and segmentation require effective multi-scale feature representation for detecting or classifying objects or regions with varying sizes.
While Convolutional Neural Networks (CNNs) have been the dominant architectures for such tasks, recently introduced Vision Transformers~(ViTs) aim to replace them as a backbone.
Similar to CNNs, ViTs build a simple multi-stage structure~(i.e., fine-to-coarse) for multi-scale representation with single-scale patches.
In this work, with a different perspective from existing Transformers, we explore multi-scale patch embedding and multi-path structure, constructing the Multi-Path Vision Transformer~(MPViT).
MPViT embeds features of the same size~(i.e., sequence length) with patches of different scales simultaneously by using overlapping convolutional patch embedding.
Tokens of different scales are then independently fed into the Transformer encoders via multiple paths and the resulting features are aggregated, enabling both fine and coarse feature representations at the same feature level.
Thanks to the diverse, multi-scale feature representations, our MPViTs scaling from tiny~(5M) to base~(73M) consistently achieve superior performance over state-of-the-art Vision Transformers on ImageNet classification, object detection, instance segmentation, and semantic segmentation.
These extensive results demonstrate that MPViT can serve as a versatile backbone network for various vision tasks.
Code will be made publicly available at \url{https://git.io/MPViT}.
\end{abstract}

\begin{figure}[t]
\begin{center}
\includegraphics[width=0.95\linewidth]{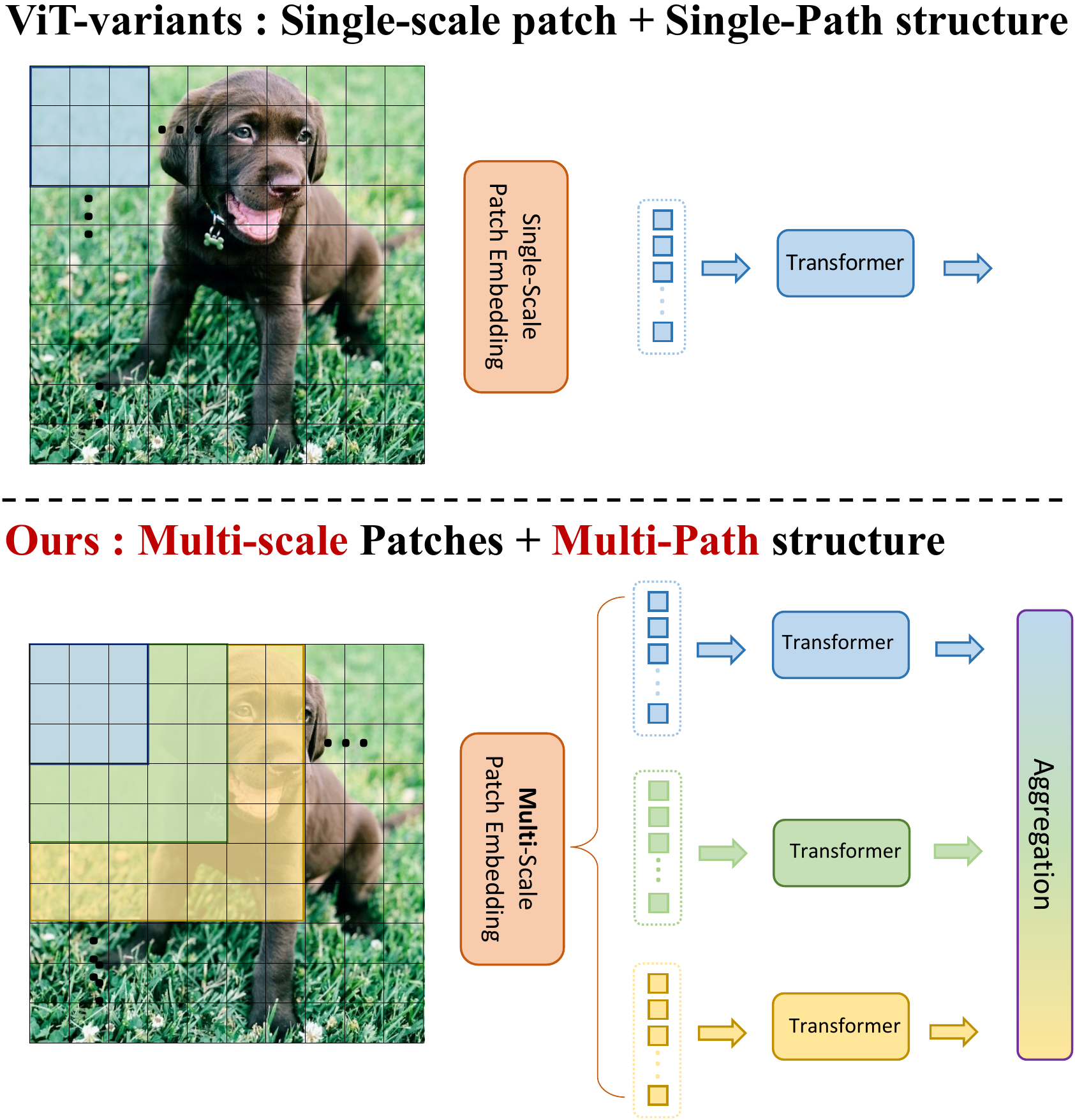}
\end{center}
\caption{\textbf{Top}: The state-of-the-art ViT-variants~\cite{wu2021cvt,liu2021swin,yang2021focal} use single-scale patches and single-path Transformer encoders. \textbf{Bottom}: MPViT uses \textbf{multi-scale patch embedding}, each embedded patch following a path to an independent Transformer encoder, allowing simulatneous representations of \textit{fine} and \textit{coarse} features.}
\label{fig:teaser}
\vspace{-0.5cm}
\end{figure}

\section{Introduction}
\label{sec:intro}

Since its introduction, the Transformer~\cite{vaswani2017attention} has had a huge impact on natural language processing~(NLP)~\cite{devlin2019bert,radford2018gpt,brown2020gpt3}. Likewise, the advent of Vision Transformer~(ViT)~\cite{dosovitskiy2021vit} has moved the computer vision community forward.
As a result, there has been a recent explosion in Transformer-based vision works, spanning tasks such as static image classification~\cite{wang2021pvt,wang2021pvtv2,liu2021swin,yang2021focal,xu2021coat,el2021xcit,touvron2021deit,touvron2021cait}, object detection~\cite{zhu2020deform-detr,carion2020detr,dai2021up}, and semantic segmentation~\cite{wang2021maxdeeplab,xie2021segformer} to temporal tasks such as video classification~\cite{bertasius2021timesformer,arnab2021vivit,Fan_2021mvit} and object tracking~\cite{meinhardt2021trackformer,wang2021tracker,chen2021xtracking}.


It is crucial for dense prediction tasks such as object detection and segmentation to represent \textit{features at multiple scales} for discriminating between objects or regions of varying sizes.
Modern CNN backbones which show better performance for dense prediction leverage multiple scales at the convolutional kernel level~\cite{szegedy2015googlenet,szegedy2017inception,lee2017wri,lee2019vovnet,gao2019res2net}, or feature level~\cite{wang2020hrnet,newell2016hourglass,lin2017fpn}.
Inception Network~\cite{szegedy2017inception} or VoVNet~\cite{lee2019vovnet} exploits multi-grained convolution kernels at the same feature level, yielding diverse receptive fields and in turn boosting detection performance.
HRNet~\cite{wang2020hrnet} represents multi-scale features by simultaneously aggregating fine and coarse features throughout the convolutional layers.

Although CNN models are widely utilized as feature extractors for dense predictions, the current state-of-the-art~(SOTA) Vision Transformers~\cite{wang2021pvt,wang2021pvtv2,zhang2021vil, liu2021swin,wu2021cvt,yang2021focal,xu2021coat,el2021xcit} have surpassed the performance of CNNs.
While the ViT-variants~\cite{wang2021pvt,wu2021cvt, zhang2021vil,liu2021swin,yang2021focal,el2021xcit} focus on how to address the quadratic complexity of self-attention when applied to dense prediction with a high-resolution, they pay less attention to building effective multi-scale representations.
For example, following conventional CNNs~\cite{simonyan2014vgg,he2016resnet}, recent Vision Transformer backbones ~\cite{wang2021pvt,zhang2021vil,liu2021swin,yang2021focal} build a \textit{simple} multi-stage structure~(\eg, fine-to-coarse structure) with \textit{single-scale} patches~(\ie, tokens). 
CoaT~\cite{xu2021coat} simultaneously represents fine and coarse features by using a co-scale mechanism allowing cross-layer attention in parallel, boosting detection performance.
However, the co-scale mechanism requires heavy computation and memory overhead as it adds extra cross-layer attention to the base models~(\eg, CoaT-Lite).
Thus, there is still room for improvement in \textit{multi-scale feature representation} for ViT architectures.

In this work, we focus on how to effectively represent \textit{multi-scale features} with Vision Transformers for dense prediction tasks.
Inspired by CNN models exploiting the multi-grained convolution kernels for multiple receptive fields~\cite{szegedy2017inception,lee2019vovnet,gao2019res2net}, we propose a \textit{multi-scale} patch embedding and \textit{multi-path} structure scheme for Transformers, called Multi-Path Vision Transformer~(MPViT).
As shown in \Fig{teaser}, the multi-scale patch embedding tokenizes the visual patches of different sizes at the same time by overlapping convolution operations, yielding features having the same sequence length~(\ie, feature resolution) after properly adjusting the padding/stride of the convolution.
Then, tokens from different scales are independently fed into Transformer encoders in parallel.
Each Transformer encoder with different-sized patches performs global self-attention. Resulting features are then aggregated, enabling both fine and coarse feature representations at the same feature level.
In the feature aggregation step, we introduce a global-to-local feature interaction (GLI) process which concatenates convolutional local features to the transformer's global features, taking advantage of both the local connectivity of convolutions and the global context of the transformer.


\begin{figure}[t]
\begin{center}
\includegraphics[width=\linewidth]{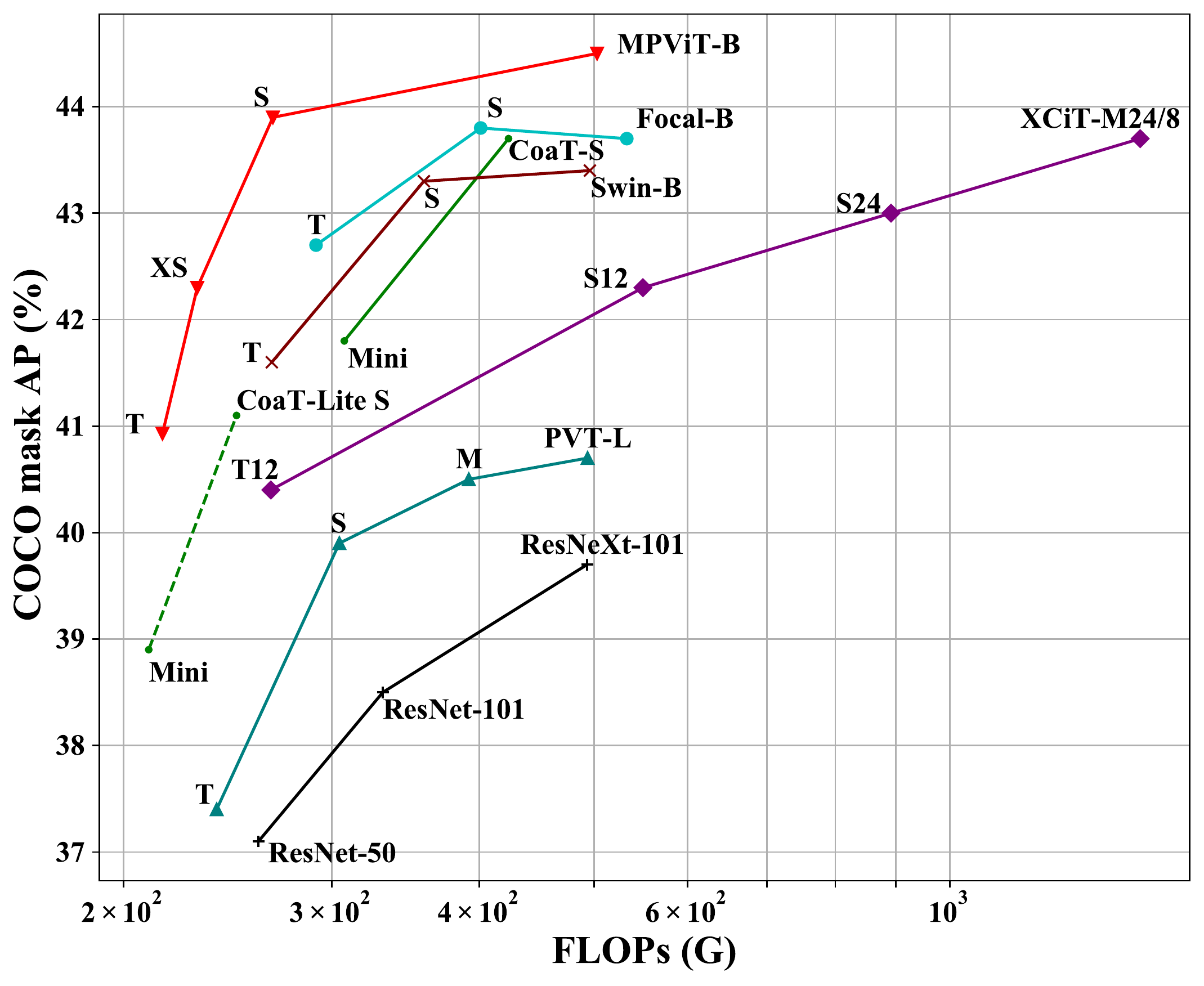}
\end{center}
\vspace{-0.1in}
\caption{\textbf{FLOPs vs. COCO mask AP} on Mask R-CNN. MPViTs outperform state-of-the-art Vision Transformers while having fewer parameters and FLOPs. B, S, XS, and T at the end of the model names denote base, small, extra-small and tiny respectively. Complete results are in \Table{det}.}
\label{fig:plot}
\vspace{-0.5cm}
\end{figure}

\begin{textblock}{42.5}(147,66.5)
      \begin{table*}[t]
  \centering
  \begingroup\setlength{\fboxsep}{0pt}
  \colorbox{white}{
  \begin{adjustbox}{width=\textwidth, center}
    \begin{tabular}{lccc}
    \textbf{Model} & \textbf{mAP} & \textbf{Param.} & \textbf{GFLOPs}\\
    \midrule
    CoaT-Lite S~\cite{xu2021coat} & 41.1 & ~~40M & ~~249\\
    CoaT-S~\cite{xu2021coat} & 43.7 & ~~42M & ~~423\\
    Swin-B~\cite{liu2021swin} & 43.4 & 107M & ~~496\\
    Focal-B~\cite{yang2021focal} & 43.7 & 110M & ~~533\\
    XCiT-M24/8~\cite{el2021xcit} & 43.7 & ~~99M & 1448\\
    \midrule
    \textbf{MPViT-S~(ours)} & \textbf{43.9} & ~~43M & ~~268\\
    \end{tabular}%
    \end{adjustbox}
    }
    \endgroup
  \label{tab:test}%
  \vspace{-0.5cm}
\end{table*}%
    \end{textblock}

Following the standard training recipe as in DeiT~\cite{touvron2021deit}, we train MPViTs on ImageNet-1K~\cite{deng2009imagenet}, which consistently achieve superior performance compared to recent SOTA Vision Transformers~\cite{liu2021swin,yang2021focal,el2021xcit,xu2021coat,wu2021cvt}.
Furthermore, We validate MPViT as a backbone on object detection and instance segmentation on the COCO dataset and semantic segmentation on the ADE20K dataset, achieving state-the-art performance.
In particular, MPViT-Small~(22M \& 4GFLOPs) surpasses the recent, and much larger, SOTA Focal-Base~\cite{yang2021focal}~(89M \& 16GFLOPs) as shown in \Fig{plot}.

To summarize, our main contributions are as follows:
\begin{itemize}
    \item We propose a multi-scale embedding with a multi-path structure for simultaneously representing fine and coarse features for dense prediction tasks.
        \vspace{-0.05in}
    \item We introduce global-to-local feature interaction~(GLI) to take advantage of both the local connectivity of convolutions and the global context of the transformer.
    \vspace{-0.05in}
    \item We provide ablation studies and qualitative analysis, analyzing the effects of different path dimensions and patch scales, discovering efficient and effective configurations.
    \vspace{-0.05in}

    \item We verify the effectiveness of MPViT as a backbone of dense prediction tasks, achieving state-of-the-art performance on ImageNet classification, COCO detection and ADE20K segmentation.
\end{itemize}

\begin{figure*}
\begin{center}
\scalebox{0.9}{
\includegraphics[width=0.97\textwidth]{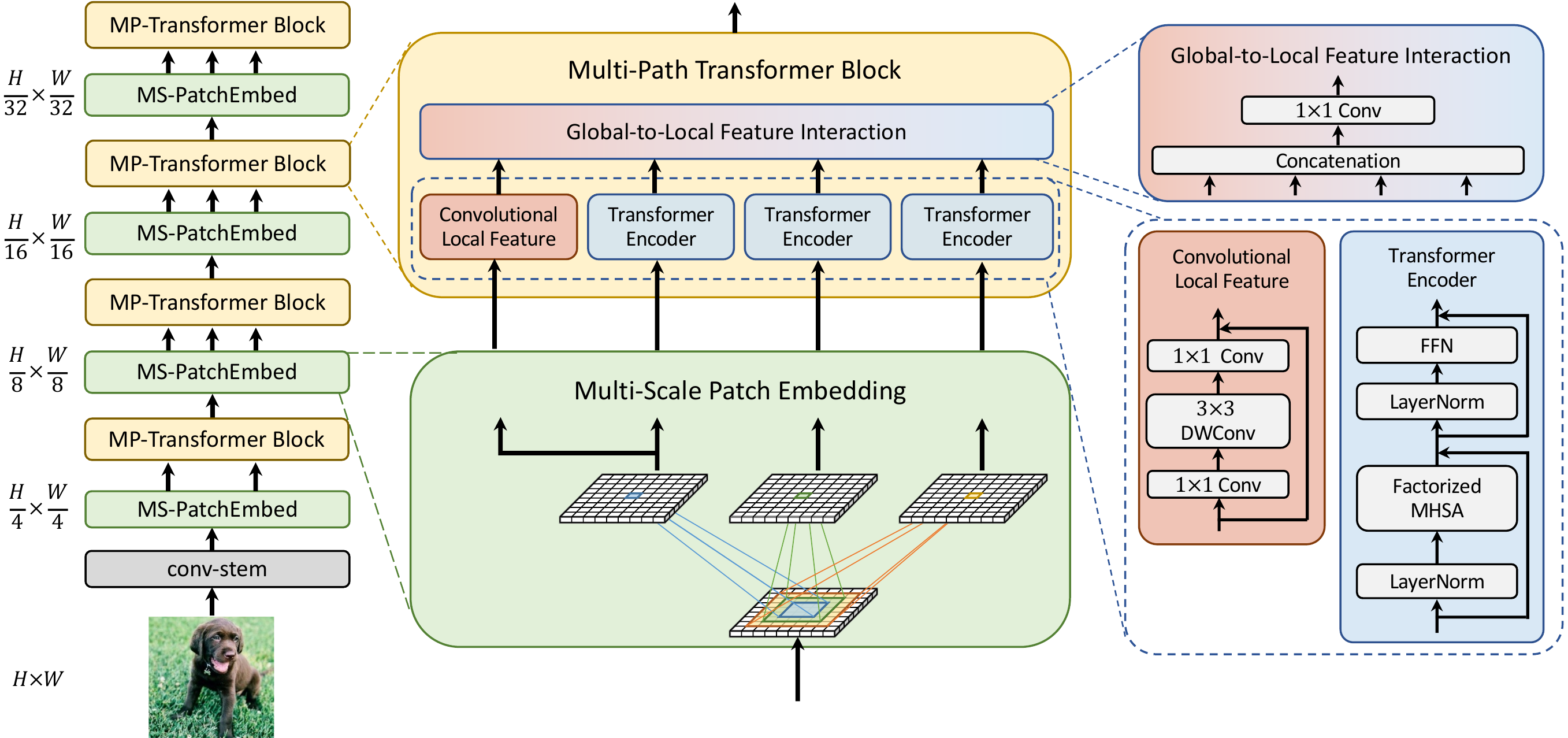}
}
\end{center}
\caption{
\textbf{Overview of Multi-Path Vision Transformer~(MPViT)}. MPViT consists of multi-scale patch embedding~(MS-PatchEmbed) and multi-path transformer~(MP-Transformer) blocks, which output features from each of the four stages for dense prediction tasks. 
Transformer encoders utilize factorized multi-head self-attention~(MHSA)~\cite{xu2021coat}. We omit the convoultional position encodings for simplicity.
}
\vspace{-0.5cm}
\label{fig:arch}
\end{figure*}
\section{Related works}
\label{sec:related}
\paragraph{Vision Transformers for dense predictions} Current SOTA Vision Transformers~\cite{wang2021pvt,zhang2021vil,liu2021swin,yang2021focal,xu2021coat,el2021xcit} have focused on reducing the quadratic complexity of self-attention when applied to dense prediction with a high-resolution.
\cite{liu2021swin,zhang2021vil,yang2021focal} constrain the attention range with fine-grained patches in local regions and combine this with sliding windows or sparse global attention.
\cite{wang2021pvt,wu2021cvt} exploit a coarse-grained global self-attention by reducing sequence length with spatial reduction~(\ie, pooling).
\cite{xu2021coat,el2021xcit} realizes linear complexity by operating the self-attention across feature channels rather than tokens.
While \cite{wang2021pvt,zhang2021vil,liu2021swin,yang2021focal} has a \textit{simple} pyramid structure~(fine-to-coarse), XCiT~\cite{el2021xcit} has a single-stage structure as ViT~\cite{dosovitskiy2021vit}.
When applied to dense prediction tasks, XCiT adds down-/up-sampling layers to extract multi-scale features after pre-training on ImageNet.
Xu \etal~\cite{xu2021coat} introduce both CoaT-Lite with a simple pyramid structure and CoaT with cross-layer attention on top of CoaT-Lite.
The cross-layer attention allows CoaT to outperform CoaT-Lite, but requires heavy memory and computation overhead, which limits scaling of the model.

\noindent
\textbf{Comparison to Concurrent work.} CrossViT~\cite{chen2021crossvit} also utilizes different patch sizes~(\eg, small and large) and dual-paths in a single-stage structure as ViT~\cite{dosovitskiy2021vit} and XCiT~\cite{el2021xcit}.
However, CrossViT's interactions between branches only occur through [CLS] tokens, while MPViT allows all patches of different scales to interact.
Also, unlike CrossViT~~(classification only), MPViT explores larger path dimensions~(\eg, over two) more generally and adopts multi-stage structure for dense predictions.
\section{Multi-Path Vision Transformer}

\subsection{Architecture}
\Fig{arch} shows the Multi-Path Vision Transformer~(MPViT) architecture.
Since our aim is to explore a powerful backbone network for dense predictions, we construct a multi-stage architecture~\cite{wang2021pvt,liu2021swin,yang2021focal} instead of a single-stage~(\textit{i.e.,} monolithic) one such as ViT~\cite{dosovitskiy2021vit} and XCiT~\cite{el2021xcit}. 
Specifically, we build a four-stage feature hierarchy for generating feature maps of different scales.
As a multi-stage architecture has features with higher resolutions, it requires inherently more computation.
Thus, we use Transformer encoders including factorized self-attention  as done in CoaT~\cite{xu2021coat} for the entire model due to its linear complexity. 
In LeViT~\cite{graham2021levit}, a convolutional stem block shows better low-level representation~(\ie, without losing salient information) than non-overlapping patch embedding.
Inspired by LeViT, given an input image with the size of $H\times W\times3$, we also adopt a stem block which consists of two $3\times3$ convolutional layers with channels of $C_2/2, C_2$ and stride of 2 which generates a feature with the size of $H/4\times W/4\times C_2$ where $C_2$ is the channel size at stage 2.
Each convolution is followed by Batch Normalization~\cite{ioffe2015bn} and a Hardswish~\cite{howard2019mobilenetv3} activation function.
From stage 2 to stage 5, we stack the proposed multi-scale patch embedding~(MS-PatchEmbed) and multi-path Transformer~(MP-Transformer) blocks in each stage.
Many works~\cite{dosovitskiy2021vit,graham2021levit,wang2021pvt,Chen_2021visformer} have proved that replacing the \texttt{[CLS]} token with a global average pooling~(GAP) of the final feature map does not affect performance,
so we also remove the \texttt{[CLS]} token and use GAP for simplicity.

\subsection{Multi-Scale Patch Embedding}\label{sec:multi-scale}
We devise a multi-scale patch embedding~(MS-PatchEmbed) layer that exploits both fine- and coarse-grained visual tokens at the same feature level.
To this end, we use convolution operations with overlapping patches, similar to CNNs~\cite{simonyan2014vgg,he2016resnet} and CvT~\cite{wu2021cvt}.
Specifically, given a 2D-reshaped output feature map~(\ie, token map) from a previous stage $\mathrm{X}_i \in \mathbb{R}^{H_{i-1} \times W_{i-1} \times C_{i-1}}$ as the input to stage $i$, we learn a function $F_{k\times k}(\cdot)$ that maps $\mathrm{X}_i$ into new tokens $F_{k\times k}(\mathrm{X}_i)$ with a channel size $C_i$, where $F(\cdot)$ is 2D convolution operation of kernel size~(\ie, patch size) $k\times k$, stride $s$ and padding $p$.
The output 2D token map $F_{k\times k}(\mathrm{X}_i) \in \mathbb{R}^{H_i \times W_i \times C_i}$ has height and width as below:
\begin{equation}
    H_i = \lfloor \frac{H_{i-1} - k + 2p}{s} + 1 \rfloor, W_i = \lfloor \frac{W_{i-1} - k + 2p}{s} + 1 \rfloor.
\end{equation}

The convolutional patch embedding layer enables us to adjust the sequence length of tokens by changing stride and padding.
\ie, it is possible to output the features of the same size~(\ie, resolution) with different patch sizes.
Thus, we form several convolutional patch embedding layers with different kernel sizes in parallel.
For example, as shown in \Fig{teaser}, we can generate various-sized visual tokens of the same sequence length with $3\times3, 5\times5, 7\times7$ patch sizes.

Since stacking consecutive convolution operations with the same channel and filter sizes enlarges receptive field~(\eg., two $3\times3$ are equivalent to $5\times5$) and requires fewer parameters~(\eg, $2\times3^2 < 5^2$), we choose consecutive $3\times3$ convolution layers in practice.
For the triple-path structure, we use three consecutive $3\times3$ convolutions with the same channel size $C'$, padding of 1 and stride of $s$ where $s$ is 2 when reducing spatial resolution otherwise 1.
Thus, given a feature $\mathrm{X}_i \in \mathbb{R}^{H_i \times W_i \times C_i}$ at stage $i$, we can get $F_{3\times3}(\mathrm{X}_i), F_{5\times5}(\mathrm{X}_i), F_{7\times7}(\mathrm{X}_i)$ features with the same size of $\frac{H_i}{s}\times \frac{C_i}{s}\times C'$.
Since MPViT has more embedding layers due to the multi-path structure, we reduce model parameters and computational overhead by adopting $3\times3$ depthwise separable convolutions~\cite{howard2017mobilenet,chollet2017xception} which consist of $3\times3$ depthwise convolution followed by $1\times1$ pointwise convolution in embedding layers.
All convolution layers are followed by Batch Normalization~\cite{ioffe2015bn} and Hardswish~\cite{howard2019mobilenetv3} activation functions.
Finally, the different sized token embedding features are separately fed into each transformer encoder.

\subsection{Global-to-Local Feature Interaction}
\label{sec:GLI}
Although self-attention in Transformers can capture long-range dependencies~(\ie, global context), it is likely to ignore structural information~\cite{islam2020much} and local relationships~\cite{lowe1999object} within each patch. 
Additionally, Transformers benefit from a \textit{shape bias}~\cite{tuli2021human}, allowing them to focus on important parts of the image.
On the contrary, CNNs can exploit local connectivity from translation invariance~\cite{kayhan2020translation,tuli2021human} -- each patch in an image is processed by the same weights.
This inductive bias encourages CNN to have a stronger dependency on \textit{texture} rather than shape when categorizing visual objects~\cite{baker2018shape}.
Thus, MPViT combines the local connectivity of CNNs with the global context transformers in a complementary manner.
To this end, we introduce a global-to-local feature interaction module that learns to interact between local and global features for enriched representations.
Specifically, to represent local feature $L_i \in \mathbb{R}^{H_i \times W_i \times C_i}$ at stage $i$, we adopt a depthwise residual bottleneck block which consists of $1\times1$ convolution, $3\times3$ depthwise convolution, and $1\times1$ convolution with the same channel size of $C_i$ and residual connection~\cite{he2016resnet}.
With the 2D-reshaped global features from each transformer $G_{i,j}\in \mathbb{R}^{H_i \times W_i \times C_i}$. Aggregation of the local and global features is performed by concatenation,
\begin{equation}
    A_{i} = \texttt{Concat}([L_i, G_{i,0}, G_{i,1}, ..., G_{i,j}])
\end{equation}
\begin{equation}
    X_{i+1} = H(A_{i}),
\end{equation}
where $j$ is the index of the path, $A_i\in \mathbb{R}^{H_i \times W_i \times (j+1)C_i}$ is the aggregated feature and $H(\cdot)$ is a function which learns to interact with features, yielding the final feature $X_{i+1}\in \mathbb{R}^{H_i \times W_i \times C_{i+1}}$ with the size of next stage channel dimension $C_{i+1}$.
We use $1\times1$ convolution with channel of $C_{i+1}$ for $H(\cdot)$.
The final feature $X_{i+1}$ is used as input for the next stage's the multi-scale patch embedding layer.

\aboverulesep=0ex
\belowrulesep=0ex


\begin{table}[t]
  \centering
 \begin{adjustbox}{width=\columnwidth, center}
\scalebox{0.85}{
\begin{tabular}{l|cccccc}
\toprule
MPViT & \#Layers & Channels & Param. & GFLOPs \\
\midrule
Tiny~(T)     & [1, 2, 4, 1] & [~~64, ~~96, 176, 216]  & ~~5.7M   & ~~1.5 \\
XSmall~(XS)    & [1, 2, 4, 1] & [~~64, 128, 192, 256]   & 10.5M  & ~~2.9  \\
Small~(S)     & [1, 3, 6, 3] & [~~64, 128, 216, 288]  & 22.8M  & ~~4.7  \\
Base~(B)    & [1, 3, 8, 3] & [128, 224, 368, 480]   & 74.8M  & 16.4  \\
\bottomrule
\end{tabular}%
}
\end{adjustbox}
  \caption{\textbf{MPViT Configurations.} MPViT models use paths [2,3,3,3] in each of the 4 stages. \#Layers and Channels denote the number of transformer encoders and the embedding dimension in each stage, respectively.
  We use 8 transformer heads in all models. The MLP expansion ratio is 2 and 4 for Tiny and other models, respectively. FLOPs are measured using $224\times224$ input image.}
  \label{tab:arch}%
  \vspace{-0.5cm}
\end{table}%

\subsection{Model Configuration}
\label{sec:config}

To alleviate the computational burden of multi path structure, we use the efficient factorized self-attention proposed in CoaT~\cite{xu2021coat}:
\begin{equation}
    \text{FactorAtt}(Q,K,V) = \frac{Q}{\sqrt{C}}(\text{softmax}(K)^\top V),
    \label{eq:factatt}
\end{equation}
where $Q,K,V\in\mathbb{R}^{N \times C}$ are linearly projected queries, keys, values and $N, C$ denote the number of tokens and the embedding dimension respectively.
To maintain comparable parameters and FLOPs, increasing the number of paths requires a reduction of the channel $C$ or the number of layers $L$~(\ie, the number of transformer encoders).
$L$ factorized self-attention layers~\cite{xu2021coat} with $N$ tokens and $h$ transformer encoder heads have a total time complexity of $\mathcal{O}(LhNC^2)$ and memory complexity of $\mathcal{O}(LhC^2+LhNC)$.
The complexities are \textit{quadratic} w.r.t. to the channel $C$ while \textit{linear} w.r.t. the number of layers $L$.
Accordingly, we expand the number of paths from single-path~(\ie, CoaT-Lite~\cite{xu2021coat} baseline) to triple-path by a reduction in $C$ rather than $L$.
We verify that reducing $C$ achieves better performance than reducing $L$ in the ablation study~(see \Table{ab1}).
As the computation cost of stage 2 is relatively high due to a higher feature resolution, we also set the number of paths to 2 at stage 2 for triple-path models.
Thus, from stage 3, triple-path models have 3 paths.

Interestingly, we also found that while triple-path and dual-path yield similar accuracy on ImageNet classification, the triple-path model shows better performance in dense prediction tasks.
This indicates that the diverse features from expanding the path dimension are useful for dense prediction tasks.
Therefore, we build MPViT models based on the triple-path structure. 
We scale-up the MPViT models from the small-scale MPViT-Tiny~(5M) corresponding to CoaT-Lite Tiny~(5M)~\cite{xu2021coat} or DeiT-Tiny(5.7M)~\cite{touvron2021deit}, to the large-scale MPViT-Base~(74M) corresponding to Swin-Base~(88M)~\cite{liu2021swin}.
All MPViT models use 8 transformer encoder heads, and the expansion ratio of the MLPs are set to 2 and 4 for Tiny and the other models, respectively.
The details of MPViTs are described in \Table{arch}.

\section{Experiments}
In this section, we evaluate the effectiveness and versatility of MPViT as a vision backbone on image classification (ImageNet-1K~\cite{deng2009imagenet}), dense predictions such as object detection and instance segmentation (COCO~\cite{lin2014coco}), and semantic segmentation (ADE20K~\cite{zhou2017ade20k}).

\subsection{ImageNet Classification}
\paragraph{Setting} We perform classification on the ImageNet-1K~\cite{deng2009imagenet} dataset. 
For fair comparison with recent works, we follow the training recipe in DeiT~\cite{touvron2021deit} as do other baseline Transformers~\cite{wang2021pvt,wang2021pvtv2,xu2021coat,liu2021swin,yang2021focal}.
We train for 300 epochs with the AdamW~\cite{loshchilov2017adamw} optimizer, a batch size of 1024, weight decay of 0.05, five warm-up epochs, and an initial learning rate of 0.001, which is scaled by a cosine decay learning rate scheduler.
We crop each image to $224\times224$ and use the same data augmentations as in~\cite{touvron2021deit,xu2021coat}.
The stochastic depth drop~\cite{huang2016dpr} is only used in the Small and Base sized models, where we set the rates to 0.05 and 0.3, respectively.
More details are described in the Appendix.

\paragraph{Results} \Table{cls} summarizes performance comparisons according to model size.
For fair comparison, we compare the models only using $224\times224$ input resolution and without distillation~\cite{touvron2021deit} or a larger resolution of $384\times384$.
MPViT models consistently outperform SOTA Vision Transformer architectures with similar parameter counts and computational complexity.
Both MPViT-XS and Small improve over the single-path baselines, CoaT-Lite Mini and Small by a large margin of 2.0\% and 1.1\%, respectively.
MPViT-Small also outperforms CoaT Small, while having about $3\times$ fewer GFLOPs.
Furthermore, MPViT-Small outperforms the larger models such as PVT-L, DeiT-B/16, and XCiT-M24/16.
MPViT-Base~(74M) achieves 84.3\%, surpassing the recent SOTA models which use more parameters such as Swin-Base~(88M) and Focal-Base~(89M).
Interestingly, the MPViT-Base outperforms XCiT-M24/16 which is trained with a more sophisticated training recipe~\cite{touvron2021cait,el2021xcit} using more epochs (400), LayerScale, and a different crop ratio.

\begin{table}[t]
  \centering
\begin{adjustbox}{width=\columnwidth, center}
  \scalebox{1.0}{
    \begin{tabular}{lcclc}
    \toprule
    Model & Param.(M) & GFLOPs & Top-1 & Reference\\
    \midrule
    DeiT-T~\cite{touvron2021deit} & ~~5.7   & ~~1.3   & 72.2 & ICML21\\
    XCiT-T12/16~\cite{el2021xcit} & ~~7.0   & ~~1.2   & 77.1 & NeurIPS21 \\
    \rowcolor{whitesmoke}CoaT-Lite T~\cite{xu2021coat} & ~~5.7   & ~~1.6   & 76.6 & ICCV21 \\
    \rowcolor{powderblue}\textbf{MPViT-T} & ~~5.8   & ~~1.6   & \textbf{78.2~(+1.6)} & \\
    \midrule
    ResNet-18~\cite{he2016resnet} & 11.7  & ~~1.8   & 69.8 &CVPR16 \\
    PVT-T~\cite{wang2021pvt} & 13.2  & ~~1.9   & 75.1 & ICCV21 \\
    XCiT-T24/16~\cite{el2021xcit} & 12.0  & ~~2.3   & 79.4 & NeurIPS21 \\
    CoaT Mi~\cite{xu2021coat} & 10.0  & ~~6.8   & 80.8 & ICCV21 \\
    \rowcolor{whitesmoke}CoaT-Lite Mi~\cite{xu2021coat} & 11.0  & ~~2.0   & 78.9 & ICCV21 \\
    \rowcolor{powderblue}\textbf{MPViT-XS} & 10.5  & ~~2.9   & \textbf{80.9~(+2.0)}& \\
    \midrule
    ResNet-50~\cite{he2016resnet} & 25.6  & ~~4.1   & 76.1 &CVPR16 \\
    PVT-S~\cite{wang2021pvt} & 24.5  & ~~3.8   & 79.8 & ICCV21 \\
    DeiT-S/16~\cite{touvron2021deit} & 22.1  & ~~4.6   & 79.9 & ICML21 \\
    Swin-T~\cite{liu2021swin} & 29.0  & ~~4.5   & 81.3 & ICCV21 \\
    CvT-13~\cite{wu2021cvt} & 20.0  & ~~4.5   & 81.6 & ICCV21 \\
    XCiT-S12/16~\cite{el2021xcit} & 26.0  & ~~4.8   & 82.0 & NeurIPS21 \\
    Focal-T~\cite{yang2021focal} & 29.1  & ~~4.9   & 82.2 & NeurIPS21 \\
    CoaT S~\cite{xu2021coat} & 22.0  & 12.6  & 82.1 & ICCV21 \\
    CrossViT-15~\cite{chen2021crossvit} & 28.2 & ~~6.1 & 82.3 & ICCV21 \\
    CvT-21~\cite{wu2021cvt} & 32.0  & ~~7.1   & 82.5 & ICCV21 \\
    CrossViT-18~\cite{chen2021crossvit} & 43.3 & ~~9.5 & 82.8 & ICCV21 \\
    \rowcolor{whitesmoke}CoaT-Lite S~\cite{xu2021coat} & 20.0  & ~~4.0   & 81.9 & ICCV21 \\
    \rowcolor{powderblue}\textbf{MPViT-S} & 22.8  & ~~4.7   & \textbf{83.0~(+1.1)}& \\
    \midrule
    ResNeXt-101~\cite{xie2017resnext} & 83.5  & 15.6  & 79.6 &CVPR17 \\
    PVT-L~\cite{wang2021pvt} & 61.4  & ~~9.8   & 81.7 & ICCV21 \\
    DeiT-B/16~\cite{touvron2021deit} & 86.6  & 17.6  & 81.8 & ICML21 \\
    XCiT-M24/16~\cite{el2021xcit} & 84.0  & 16.2  & 82.7 & NeurIPS21 \\
    Swin-B~\cite{liu2021swin} & 88.0  & 15.4  & 83.3 & ICCV21 \\
    XCiT-S12/8~\cite{el2021xcit} & 26.0  & 18.9  & 83.4 & NeurIPS21 \\
    Focal-B~\cite{yang2021focal} & 89.8  & 16.0  & 83.8  & NeurIPS21 \\
    \rowcolor{powderblue}\textbf{MPViT-B} & 74.8  & 16.4  & \textbf{84.3}&  \\
    \bottomrule
    \end{tabular}%
    }
    \end{adjustbox}
 \caption{\textbf{ImageNet-1K classification.} These models are trained with $224\times224$ resolution. For fair comparison, we do not include models that are distilled~\cite{touvron2021deit} or use $384\times384$ resolution. Note that CoaT-Lite~\cite{xu2021coat} models are our single-path baselines.}
  \label{tab:cls}%
\end{table}%

\begin{table*}[htbp]
  \centering
      \begin{adjustbox}{max width=\textwidth}\
\begin{tabular}{lll|cccccc|cccccc}
    \toprule
    \multirow{2}[2]{*}{Backbone} & \multirow{2}[2]{*}{Params. (M)} & \multirow{2}[2]{*}{GFLOPs}  & \multicolumn{6}{c|}{Mask R-CNN $3\times$ schedule + MS} & \multicolumn{6}{c}{RetinaNet $3\times$ schedule + MS} \\
          &       &             & $AP^b$ & $AP^b_{50}$ & $AP^b_{75}$ & $AP^m$ & $AP^m_{50}$ & $AP^m_{75}$ & $AP^b$ & $AP^b_{50}$ & $AP^b_{75}$ & $AP^b_S$ & $AP^b_M$ & $AP^b_L$ \\
    \midrule
    XCiT-T12/16~\cite{el2021xcit} & ~~26    & ~~200    & 42.7  & 64.3  & 46.4  & 38.5  & 61.2  & 41.1  & -     & -     & -     & -     & -     & - \\
    XCiT-T12/8~\cite{el2021xcit} & ~~26    & ~~266     & 44.5  & 66.4  & 48.8  & 40.4  & 63.5  & 43.3  & -     & -     & -     & -     & -     & - \\
    \rowcolor{powderblue} \textbf{MPViT-T} & ~~28 (17) & ~~216 (196)  & \textbf{44.8}  & 66.9  & 49.2  & \textbf{41.0}  & 64.2  & 44.1  & \textbf{44.4}  & 65.5  & 47.4  & 29.9  & 48.3  & 56.1  \\
    \midrule
    PVT-T~\cite{wang2021pvt} & ~~33 (23) & ~~240 (221)   & 39.8  & 62.2  & 43.0  & 37.4  & 59.3  & 39.9  & 39.4  & 59.8  & 42.0  & 25.5  & 42.0  & 52.1  \\
    CoaT Mini~\cite{xu2021coat} & ~~30    & ~~307      & 46.5  & 67.9  & 50.7  & 41.8  & 65.3  & 44.8  & -     & -     & -     & -     & -     & - \\
    \rowcolor{whitesmoke}CoaT-Lite Mini~\cite{xu2021coat} & ~~31    & ~~210     & 42.9  & 64.7  & 46.7  & 38.9  & 61.6  & 41.7  & -     & -     & -     & -     & -     & - \\
    \rowcolor{powderblue} \textbf{MPViT-XS} & ~~30 (20) & ~~231 (211)   & \textbf{46.6}  & 68.5  & 51.1  & \textbf{42.3}  & 65.8  & 45.8  & \textbf{46.1}  & 67.4  & 49.3  & 31.4  & 50.2  & 58.4  \\
    \midrule
    PVT-S~\cite{wang2021pvt} & ~~44 (34) & ~~305 (226)   & 43.0  & 65.3  & 46.9  & 39.9  & 62.5  & 42.8  & 42.2  & 62.7  & 45.0  & 26.2  & 45.2  & 57.2  \\
    XCiT-S12/16~\cite{el2021xcit} & ~~44    & ~~285     & 45.3  & 67.1  & 49.5  & 40.8  & 64.0  & 43.8  & -     & -     & -     & -     & -     & - \\
    Swin-T~\cite{liu2021swin} & ~~48 (39) & ~~267 (245)   & 46.0  & 68.1  & 50.3  & 41.6  & 65.1  & 44.9  & 45.0  & 65.9  & 48.4  & 29.7  & 48.9  & 58.1  \\
    XCiT-S12/8~\cite{el2021xcit} & ~~43    & ~~550      & 47.0  & 68.9  & 51.7  & 42.3  & 66.0  & 45.4  & -     & -     & -     & -     & -     & - \\
    Focal-T~\cite{yang2021focal} & ~~49 (39) & ~~291 (265)      & 47.2  & 69.4  & 51.9  & 42.7  & 66.5  & 45.9  & 45.5  & 66.3  & 48.8  & 31.2  & 49.2  & 58.7  \\
    CoaT S~\cite{xu2021coat} & ~~42    & ~~423      & \textbf{49.0}  & 70.2  & 53.8  & 43.7  & 67.5  & 47.1  & -     & -     & -     & -     & -     & - \\
    \rowcolor{whitesmoke}CoaT-Lite S~\cite{xu2021coat} & ~~40    & ~~249      & 45.7  & 67.1  & 49.8  & 41.1  & 64.1  & 44.0  & -     & -     & -     & -     & -     & - \\
    \rowcolor{powderblue} \textbf{MPViT-S} & ~~43 (32) & ~~268 (248)    & 48.4  & 70.5  & 52.6  & \textbf{43.9}  & 67.6  & 47.5  & \textbf{47.6}  & 68.7  & 51.3  & 32.1  & 51.9  & 61.2  \\
    \midrule
    PVT-M~\cite{wang2021pvt} & ~~64 (54) & ~~392 (283)    & 44.2  & 66.0  & 48.2  & 40.5  & 63.1  & 43.5  & 43.2  & 63.8  & 46.1  & 27.3  & 46.3  & 59.9  \\
    PVT-L~\cite{wang2021pvt} & ~~81 (71) & ~~494 (345)    & 44.5  & 66.0  & 48.3  & 40.7  & 63.4  & 43.7  & 43.4  & 63.6  & 46.1  & 26.1  & 46.0  & 59.5  \\
    XCiT-M24/16~\cite{el2021xcit} & 101   & ~~523      & 46.7  & 68.2  & 51.1  & 42.0  & 65.5  & 44.9  & -     & -     & -     & -     & -     & - \\
    XCiT-S24/8~\cite{el2021xcit} & ~~65 & ~~892      & 48.1  & 69.5  & 53.0  & 43.0  & 66.5  & 46.1  & -  & -  & -  & -  & -  & -  \\
    XCiT-M24/8~\cite{el2021xcit} & ~~99 & 1448       & 48.5  & 70.3  & 53.4  & 43.7  & 67.5  & 46.9  & -  & -  & -  & -  & -  & -  \\
    Swin-S~\cite{liu2021swin} & ~~69 (60) & ~~359 (335)   & 48.5  & 70.2  & 53.5  & 43.3  & 67.3  & 46.6  & 46.4  & 67.0  & 50.1  & 31.0  & 50.1  & 60.3  \\
    Swin-B~\cite{liu2021swin} & 107 (98) & ~~496 (477)   & 48.5  & 69.8  & 53.2  & 43.4  & 66.8  & 49.6  & 45.8  & 66.4  & 49.1  & 29.9  & 49.4  & 60.3  \\
    Focal-S~\cite{yang2021focal} & ~~71 (62) & ~~401 (367)      & 48.8  & 70.5  & 53.6  & 43.8  & 67.7  & 47.2  & 47.3  & 67.8  & 51.0  & 31.6  & 50.9  & 61.1  \\
    Focal-B~\cite{yang2021focal} & 110 (101) & ~~533 (514)      & 49.0  & 70.1  & 53.6  & 43.7  & 67.6  & 47.0  & 46.9  & 67.8  & 50.3  & 31.9  & 50.3  & 61.5  \\
    \rowcolor{powderblue} \textbf{MPViT-B} & ~~95 (85) & ~~503 (482)  & \textbf{49.5}      & 70.9      & 54.0       &\textbf{44.5}       &68.3     &48.3      &\textbf{48.3}       &69.5     &51.9       &32.3       &52.2       &62.3  \\
    \bottomrule
    \end{tabular}%
    \end{adjustbox}
    \caption{\textbf{COCO detection and instance segmentation} with RetinaNet~\cite{lin2017retinanet} and Mask R-CNN~\cite{he2017mask}. Models are trained for $3\times$ schedule~\cite{wu2019detectron2} with multi-scale training inputs~(MS)~\cite{liu2021swin,sun2021sparsercnn}. 
    All backbones are pretrained on ImageNet-1K. We omit models pretrained on larger-datasets~(\eg, ImageNet-21K). 
    Mask R-CNN's parameters/FLOPs  are followed by RetinaNet in parentheses.}
  \label{tab:det}%
\end{table*}%

\subsection{Object Detection and Instance Segmentation}
\paragraph{Setting} We validate MPViT as an effective feature extractor for object detection and instance segmentation with RetinaNet~\cite{lin2017retinanet} and Mask R-CNN~\cite{he2017mask}, respectively.
We benchmark our models on the COCO~\cite{lin2014coco} dataset.
We pre-train the backbones on the ImageNet-1K and plug the pretrained backbones into RetinaNet and Mask R-CNN.
Following common settings~\cite{he2017mask,wu2019detectron2} and the training recipe of Swin-Transformer~\cite{liu2021swin}, we train models for $3\times$ schedule~(36 epochs)~\cite{wu2019detectron2} with a multi-scale training strategy~\cite{liu2021swin,sun2021sparsercnn,carion2020detr}. 
We use AdamW~\cite{loshchilov2017adamw} optimizer with an initial learning rate of 0.0001 and weight decay of 0.05.
We implement models based on the \texttt{detectron2}~\cite{wu2019detectron2} library.
More details are described in the Appendix.

\paragraph{Results}
\Table{det} shows MPViT-models consistently outperform recent, comparably sized SOTA Transformers on both object detection and instance segmentation.
For RetinaNet, MPViT-S achieves 47.6\%, which improves over Swin-T~\cite{liu2021swin} and Focal-T~\cite{yang2021focal}, by large margins of over 2.1 - 2.6\%.
Interestingly, MPViT-S~(32M) shows superior performance compared to the much larger Swin-S~(59M) / B~(98M) and Focal-S~(61M) / B~(100M), which have higher classification accuracies in Table~\ref{tab:cls}.
These results demonstrate the proposed multi-scale patch embedding and multi-path structure can represent more diverse multi-scale features than simpler multi-scale structured models for object detection, which requires scale-invariance.
Notably, Swin-B and Focal-B show a performance drop compared to Swin-S and Focal-S, while MPViT-B improves over MPViT-S, showing MPViT scales well to large models.

For Mask R-CNN, MPViT-XS and MPViT-S outperform the single-path baselines CoaT~\cite{xu2021coat}-Lite Mini and Small by significant margins.
Compared to CoaT which adds parallel blocks to CoaT-Lite with additional cross-layer attention, MPViT-XS improves over CoaT Mini, while MPViT-S shows lower box $AP^b$ but higher mask $AP^m$.
We note that although CoaT-S and MPViT-S show comparable performance, MPViT-S requires much less computation.
This result suggests that MPViT can \textit{efficiently} represent multi-scale features without the additional cross-layer attention of CoaT.
Notably, the mask AP~(43.9\%) of MPViT-S is higher than those of larger models such as XCiT-M24/8 or Focal-B, while having much less FLOPs.

\begin{table}[t]
  \centering
  \begin{adjustbox}{width=0.82\columnwidth, center}
  \scalebox{0.8}{
    \begin{tabular}{lccc}
    \toprule
    Backbone & Params. & GFLOPs    & mIoU \\
    \midrule
    Swin-T~\cite{liu2021swin} & ~~59M   & ~~945     & 44.5 \\
    Focal-T~\cite{yang2021focal} & ~~62M   & ~~998      & 45.8 \\
    XCiT-S12/16~\cite{el2021xcit} & ~~54M   & ~~966     & 45.9 \\
    XCiT-S12/8~\cite{el2021xcit} & ~~53M   & 1237    & 46.6 \\
    \rowcolor{powderblue}\textbf{MPViT-S} & ~~52M & ~~943  & \textbf{48.3} \\
    \midrule
    XCiT-S24/16~\cite{el2021xcit} & ~~76M   & 1053    & 46.9 \\
    Swin-S~\cite{liu2021swin} & ~~81M   & 1038    & 47.6 \\
    XCiT-M24/16~\cite{el2021xcit} & 112M  & 1213    & 47.6 \\
    Focal-S~\cite{yang2021focal} & ~~85M   & 1130     & 48.0 \\
    Swin-B~\cite{liu2021swin} & 121M  & 1841    & 48.1 \\
    XCiT-S24/8~\cite{el2021xcit} & ~~74M   & 1587     & 48.1 \\
    XCiT-M24/8~\cite{el2021xcit} & 110M  & 2161     & 48.4 \\
    Focal-B~\cite{yang2021focal} & 126M  & 1354     & 49.0 \\
    \rowcolor{powderblue}\textbf{MPViT-B} &   105M   &   1186          & \textbf{50.3} \\
    \bottomrule
    \end{tabular}%
    }
\end{adjustbox}
  \caption{\textbf{ADE20k semantic segmentation} results using UperNet~\cite{xiao2018upernet}. 
  For fair comparison, We do not include models that are pre-trained on larger datasets~(\textit{i.e.,} ImageNet-21K).}
  \label{tab:seg}%
  \vspace{-0.5cm}
\end{table}%

\begin{table*}[t]
  \centering
  \begin{adjustbox}{width=\textwidth, center}
  \scalebox{0.95}{
    \begin{tabular}{llcccclll}
    \toprule
    Path & Spec  & Param. & GFLOPs & Memory & img/sec &  Top-1 & AP\textsuperscript{box}  & AP\textsuperscript{mask} \\
    \midrule
    Single & [1,1,1,1]P\_[2,2,2,2]L\_[64, 128, 320, 512]C & 11.0M  & 1.9   & 9216  & 1195  & 78.9  & 40.2  & 37.3 \\ \midrule
    (a) Dual & [2,2,2,2]P\_[\textcolor{RedOrange}{1,2,4,1}]L\_[\textcolor{DarkOrchid}{64, 128, 256, 320}]C & 10.9M  & 2.6   & 6054  & ~~945   & 80.7\scriptsize{\textcolor{Green}{+1.8}}  & 42.6\scriptsize{\textcolor{Green}{+2.4}}  & 39.1\scriptsize{\textcolor{Green}{+1.8}} \\
    (b) Triple & [2,3,3,3]P\_[1,1,2,1]L\_[\textcolor{DarkOrchid}{64, 128, 256, 320}]C & 10.8M  & 2.3   & 6000   & 1080  & 79.8\scriptsize{\textcolor{Green}{+0.9}}  & 41.4\scriptsize{\textcolor{Green}{+1.2}}  & 38.0\scriptsize{\textcolor{Green}{+0.7}} \\    
    (c) Triple & [2,3,3,3]P\_[\textcolor{RedOrange}{1,2,4,1}]L\_[64, 128, 192, 256]C & 10.1M  & 2.7   & 5954 & ~~803   & 80.5\scriptsize{\textcolor{Green}{+1.6}}  & \textbf{43.0}\scriptsize{\textcolor{Green}{+2.8}}  & \textbf{39.4}\scriptsize{\textcolor{Green}{+2.1}} \\
    (d) Quad & [2,4,4,4]P\_[\textcolor{RedOrange}{1,2,4,1}]L\_[64, ~~96, 176, 224]C & 10.5M  & 2.6   & 5990 & ~~709   & 80.5\scriptsize{\textcolor{Green}{+1.6}}  & 42.4\scriptsize{\textcolor{Green}{+2.2}}  & 38.8\scriptsize{\textcolor{Green}{+1.5}} \\    

    \bottomrule
    \end{tabular}%
    }
    \end{adjustbox}
      \caption{\textbf{Exploring the path dimension.} Spec means [\#path\_per\_stage]P, [\#layer\_per\_stage]L and [dimension\_per\_stage]C. We measure inference throughput and peak GPU memory usage on V100 GPU with batch size of 256. Note that the single-path is CoaT-Lite Mini~\cite{xu2021coat}.}
  \label{tab:ab1}%
  \vspace{-0.3cm}
\end{table*}%


\begin{table}[t]
  \centering
    \begin{adjustbox}{width=\columnwidth, center}
    \begin{tabular}{lcccc}
    \toprule
    Path & Param. & GFLOPs & Top-1 & AP\textsuperscript{b}/AP\textsuperscript{m} \\
    \midrule
    Single~(CoaT-Lite Mini)  & 11.01M & 1.99  & 78.9  & 40.2 / 37.3 \\ \midrule
    + Triple~(p=[3,5,7], parallel)& 10.18M & 2.78  & 80.3  & 41.7 / 38.4 \\
    + Triple~(p=[3,3,3], series)& 10.15M & 2.67  & \textbf{80.5}  & \textbf{43.0} / \textbf{39.4} \\ \midrule
    + GLI~(Sum)   & 10.13M & 2.82  & 80.3  & 43.0 / 39.5 \\
    + GLI~(Concat.)   & 10.57M & 2.97  & \textbf{80.8}  & \textbf{43.3} / \textbf{39.7} \\
    \bottomrule
    \end{tabular}%
    \end{adjustbox}
  \caption{\textbf{Component Analysis.}}
  \label{tab:ab2}%
  \vspace{-0.3cm}
\end{table}%

\subsection{Semantic segmentation}
\paragraph{Setting} We further evaluate the capability of MPViT for semantic segmentation on the ADE20K~\cite{zhou2017ade20k} dataset.
We deploy UperNet~\cite{xiao2018upernet} as a segmentation method and integrate the ImageNet-1k pre-trained MPViTs into the UperNet.
Following~\cite{liu2021swin,el2021xcit}, for fair comparison, we train models for 160k iterations with a batch size of 16, the AdamW~\cite{loshchilov2017adamw} optimizer, a learning rate of 6e-5, and a weight decay of 0.01.
We report the performance using the standard single-scale protocol.
We implement MPViTs using \texttt{mmseg}~\cite{mmseg2020} library.
More details are described in the Appendix.

\paragraph{Results} As shown in \Table{seg}, our MPViT models consistently outperform recent SOTA architectures of similar size.
MPViT-S achieves higher performance~(48.3\%) over other Swin-T, Focal-T and XCiT-S12/16 by large margins of  +3.8\%, +2.5\%, and +2.4\%.
Interestingly, MPViT-S also surpasses much larger models, \eg, Swin-S/B, XCiT-S24/16, -M24/16, -S24/8, and Focal-S.
Furthermore, MPViT-B outperforms the recent (and larger) SOTA Transformer, Focal-B~\cite{yang2021focal}.
These results demonstrate the diverse feature representation power of MPViT, which stems from its multi-scale embedding and multi-path structure, makes MPViT effective on pixel-wise dense prediction tasks.

\subsection{Ablation study}
We conduct ablation studies on each component of MPViT-XS to investigate the effectiveness of the proposed multi-path structure on image classification and object detection with Mask R-CNN~\cite{he2017mask} using $1\times$ schedule~\cite{he2017mask} and single-scale input.

\paragraph{Exploring path dimension}
\label{sec:path}
We investigate the effect of differing path dimensions, and how the path dimension could be effectively extended in \Table{ab1}.
We conduct experiments using various metrics such as model size~(\ie, model parameter), computation cost~(GFLOPs), GPU peak memory, and GPU throughput~(img/sec).
We use Coat-Lite Mini~\cite{xu2021coat} as a single-path baseline because it also leverages the same factorized self-attention mechanism as MPViT.
For a fair comparison with the baseline, we do not use a stem block, stochastic depth drop path, and the convolutional local features introduced in Sec.~\ref{sec:GLI}.
For dual-path, higher feature resolution at stage 2 requires more computation, so we decrease the number of layers~$L$ (\ie, the number of transformer encoders).
At stage 5, a higher embedding dimension results in a larger model size, thus we also reduce $L$ and the embedding dimension~$C$, increasing $L$ at stage 4 instead.
As multiple paths lead to higher computation cost, we curtail $C$ at stages 3 and 4 to compensate.
As a result, dual-path~(a) in \Table{ab1} improves over the single-path while having a similar model size and slightly higher FLOPs.

When expanding dual-path to triple-path, we ablate the embedding dimension~$C$ and the number of layers~$L$, respectively.
For the embedding dimension of (b) in \Table{ab1}, we maintain $C$ but reduce $L$ to maintain a similar model size and FLOPs, which leads to worse accuracy than the dual-path.
Conversely, when we decrease $C$ and maintain $L$, (c) achieves similar classification accuracy but higher detection performance than the dual-path.
Lastly, we further expand the path to quad-path (d), keeping $L$ and reducing $C$.
The quad-path achieves similar classification accuracy, but detection performance is not better than the triple-path of (c).
These results teach us three lessons : \textit{i}) the number of layers~(\ie, \textit{deeper}) is more important than the embedding dimension~(\ie, \textit{wider}), which means \textit{deeper and thinner} structure is better in terms of performance. 
\textit{ii}) the multi-grained token embedding and multi-path structure can provide object detectors with richer feature representations.
\textit{iii}) Under the constraint of the same model size and FLOPs, triple-path is the best choice. 

We note that our strategy of expanding the path dimension does not increase the memory burden as shown in \Table{ab1}.
dual-path~(a) and triple-path~(b,c) consume less memory than the single-path.
Also, (a) and (b) consume more memory than (c) because (a) and (b) have bigger $C$ at stages 3 and 4.
This is because $C$~(quadratic) is a bigger factor in memory usage than $L$~(linear) as described in sec.\ref{sec:config}.
Therefore, our strategy of reducing the embedding dimension and expanding the path dimension and layers~(deeper) leads to a \textit{memory-efficient} model.
However, the growth of the total number of layers due to multi-path structure decreases inference speed as compared the single-path.
This issue is addressed in detail in sec.~\ref{sec:limitation}.

\vspace{-0.05in}
\paragraph{Multi-Scale Embedding}
In \Table{ab2}, we investigate the effects of patch size and structure in the multi-scale embedding, as outlined in Section~\ref{sec:multi-scale}.
We use three convolution layers in \textit{parallel} with the same stride of 2 and patch sizes of 3, 5, and 7, respectively.
\ie, each path embedding layer operates independently using the previous input feature.
For parameter efficiency, we also use three convolution layers in \textit{series} with the same kernel size of 3 and strides of 2, 1, 1, respectively.
We note that the latter has equivalent receptive fields~(\eg, 3, 5, 7) as shown in \Fig{arch}.
The \textit{series} version improves over \textit{parallel} while reducing the model size and FLOPs. 
Intuitively, this performance gain likely comes from the fact that the series version actually contains small 3 layer CNNs with non-linearities, which allows for more complex representations.

\vspace{-0.05in}
\paragraph{Global-to-Local feature Interaction}
We experiment with different aggregation schemes in the GLI module, which aggregates convolutional local feature and the global transformer features, we test two types of operations: \textit{addition} and \textit{concatenation}.
As shown in \Table{ab2}, the sum operation shows no performance gain while concatenation shows improvement on both classification and detection tasks. Intuitively, summing features before the $1\times1$ convolution naively mixes the features, while concatenation preserves them, allowing the $1\times1$ convolution to learn more complex interactions between the features.
This result demonstrates that the GLI module effectively learns to interact between local and global features for enriching representations.
\begin{table}[t]
  \centering
  \begin{adjustbox}{width=\columnwidth, center}
    \begin{tabular}{lcccccc}
    \toprule
    Model & Top-1 & AP\textsuperscript{b}/AP\textsuperscript{m}& Param. & GFLOPs & Mem. & img/s \\
    \midrule
    Swin-T & 81.3  & 46.0/41.6 & 28M   & ~~4.5   & 10.4 & 1021 \\
    Focal-T & 82.2  & 47.2/42.7 & 29M & ~~4.9   & 19.3 & ~~400 \\
    XCiT-S12/16 & 82.0    & 45.3/40.8 & 26M   & ~~4.8   & ~~\textbf{6.5}  & \textbf{1181} \\
    XCiT-S12/8 & \textbf{83.4} & 47.0/42.3 & 26M   & 18.9 & 10.5 & ~~318 \\
    CoaT S & 82.1  & \textbf{49.0}/43.7 & 22M   & 12.6  & 13.3 & ~~121 \\
    CoaT-Lite S & 82.0    & 45.7/41.1 & \textbf{20M}   & ~~\textbf{4.0}   & ~~9.8  & ~~688 \\
    \textbf{MPViT-S} & 83.0    & 48.4/\textbf{43.9} & 22M   & ~~4.7   & ~~7.2  & ~~546 \\
    \bottomrule
    \end{tabular}%
    \end{adjustbox}
  \caption{\textbf{Model Capacity Analysis.} We measure inference throughput and peak GPU memory usage (GB) for MPViT-S and comparable models. All models are tested on V100 GPU with batch size of 256 and $224\times224$ resolution.}
  \label{tab:ab3}%
  \vspace{-0.3cm}
\end{table}%

\section{Discussion}
\vspace{-0.1in}
\paragraph{Model Capacity Analysis}
\label{sec:model}
Measuring actual GPU throughput and memory usage, we analyze the model capacity of MPViT-S, comparing with recent SOTA Transformers~\cite{xu2021coat,liu2021swin,yang2021focal,el2021xcit} in \Table{ab3}.
We test all models on the same Nvidia V100 GPU with a batch size of 256.
Although CoaT Small~\cite{xu2021coat} achieves the best detection performance thanks its additional cross-layer attention, it exhibits heavier memory usage and GPU computation than CoaT-Lite Small with a simple multi-stage structure similar to Swin-T~\cite{liu2021swin} and Focal-T~\cite{yang2021focal}.
Compared to CoaT Small, MPViT-S consumes much less memory and runs $4\times$ faster with comparable detection performance, which means MPViT can perform efficiently and its multi-scale representations are effective without the additional cross layer attention of CoaT.
Moreover, CoaT has limitations in scaling up models due to its exhaustive memory usage, but MPViT can scale to larger models.
For XCiT~\cite{el2021xcit} having single-stage structure, XCiT-S12/16~(16x16 patches : scale 4) shows faster speed and less memory usage, while XCiT-S12/8 requires more computation and memory than MPViT-S due to its higher feature resolution.
We note that XCiT-S12/8 shows higher classification accuracy~(83.4\%) than MPViT-S~(83.0\%), whereas detection performance is the opposite~(47.0 vs. 48.4).
This result demonstrates that \textit{for dense prediction tasks}, the mutli-scale embedding and multi-path structure of MPViT is both more efficient and effective than the single-stage structure of XCiT equipped with additional up-/down-sampling layers.
MPViT also has a relatively smaller memory footprint than most models.

\begin{figure}[t]
\begin{center}
\includegraphics[width=\linewidth]{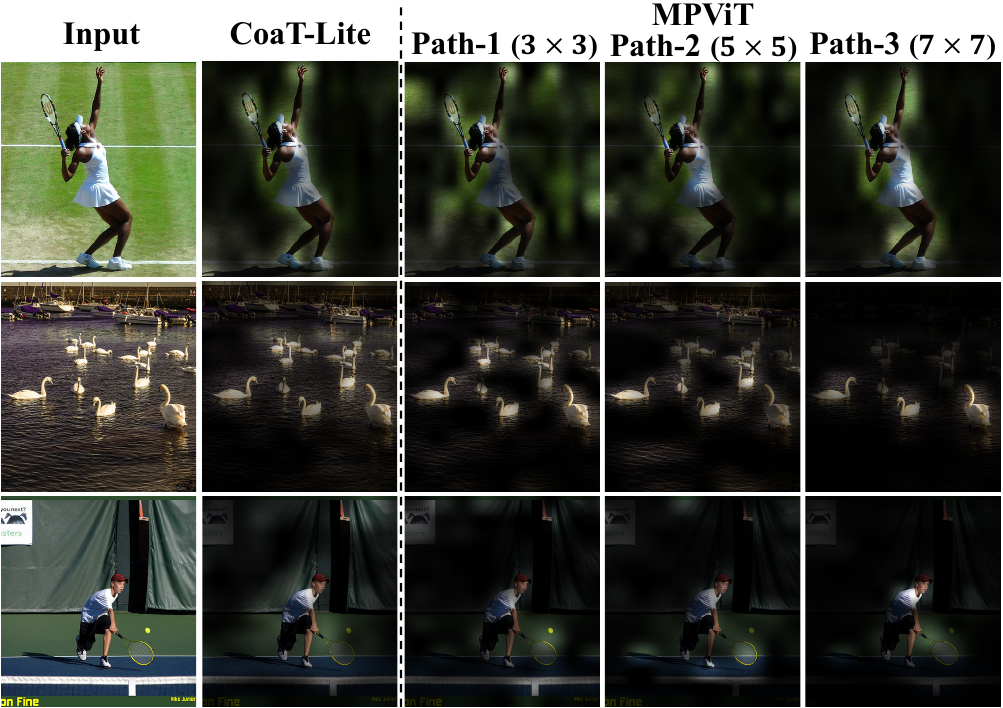}
\end{center}
\vspace{-0.05in}
\caption{\textbf{Attention maps} generated by CoaT-Lite and MPViT at stage4. MPViT has a triple-path structure with patches of various sizes~(\eg, $3\times3, 5\times5, 7\times7$), leading to fine and coarse features. See Appendix for more visualization results.}
\label{fig:attention}
\vspace{-0.2in}
\end{figure}


\paragraph{Qualitative Analysis}
\label{sec:vis}
In \Fig{attention}, we visualize the attention maps, comparing the triple-path (c in \Table{ab1}) with the single-path~(CoaT-Lite Mini).
Since the triple-path embeds different patch sizes, we visualize attention maps for each path.
The attention maps from CoaT-Lite and path-1 have similar patch sizes and show similar attention maps.
Interestingly, we observe that attention maps from path-3, which attends to larger patches with higher-level representations, are more object centric, precisely capturing the extents of the objects, as shown in the rightmost column of \Fig{attention}. However, at the same time, path-3 suppresses small objects and noise. Contrarily, path-1 attends to small objects due to fine patches, but does not precisely capture large-object boundaries due to its usage of low-level representations. 
This is especially apparent in the 3rd-row of \Fig{attention}, where path-1 captures a smaller ball, while path-3 attends to a larger person.  These results demonstrate that combining fine and coarse features via a multi-path structure can capture objects of varying scales in the given visual inputs.


\paragraph{Limitation and Future works}
\label{sec:limitation}
The extensive experimental results have demonstrated that MPViT significantly outperforms current SOTA Vision Transformers not only on image-level prediction, but also on dense predictions tasks. A possible limitation of our MPViT model is the latency at inference time. As shown in \Table{ab3}, the inference time of MPViT-S is slower than Swin-T and XCiT-S12/16.
We hypothesize that the multi-path structure leads to suboptimal GPU utilization as similar observations have been made for grouped convolution ~\cite{xie2017resnext,ma2018shufflenetv2,lee2019vovnet}~(\eg, GPU context switching, kernel synchronization, etc.). 
To alleviate this issue, we are currently working on an efficient implementation of our model to speed-up the inference of MPViT, which integrates the features with different scales into one tensor and then performing multi-head self attention with the tensor. This will improve parallelization and GPU utilization. Moreover, to strike the balanced tradeoff between accuracy/speed, we will further consider path dimensions in a compound scaling strategy~\cite{tan2019efficientnet,dollar2021fast}, to consider the optimal combination of depths, widths, and resolutions when increasing/decreasing the model capacity.


\section{Acknowledgement}
This work was supported by Institute of Information \& Communications Technology Planning \& Evaluation (IITP) grant funded by the Korean government (MSIT) (No. 2020-0-00004, Development of Previsional Intelligence based on Long-term Visual Memory Network and No. 2014-3-00123, Development of High Performance Visual BigData Discovery Platform for Large-Scale Realtime Data Analysis).
\appendix




\section{Appendix}
In this appendix, Section~\ref{sec:setting} first describe the training details of our experiments for ImageNet classification, COCO detection/instance segmentation, and ADE20K semantic segmentation.
Second, in Section~\ref{sec:ex}, we show further experimental analyses for ImageNet classification and COCO object detection.
Finally, in Section~\ref{sec:vis_supp}, we provide more qualitative analysis on the learned attention maps and failure cases.

\subsection{Detailed Experimental Settings}
\label{sec:setting}
\paragraph{ImageNet classification} Following the training recipe as in CoaT~\cite{xu2021coat} and DeiT~\cite{devlin2019bert}, we perform the same data augmentations such as MixUP~\cite{inoue2018mixup}, CutMix~\cite{yun2019cutmix}, random erasing~\cite{zhong2020random}, repeated augmentation~\cite{hoffer2020augment}, and label smoothing~\cite{szegedy2016rethinking}.
We train MPViTs for 300 epochs with the AdamW~\cite{loshchilov2017adamw} optimizer, a batch size of 1024, weight decay of 0.05, five warm-up epochs, and an initial learning rate of 0.001, which is scaled by a cosine decay learning rate scheduler.
We implement MPViTs based on CoaT official code~\footnote{\url{https://github.com/mlpc-ucsd/CoaT}} and \texttt{timm} library~\cite{rw2019timm}.

\begin{table}[t]
  \centering
\begin{adjustbox}{width=\columnwidth, center}
  \scalebox{1.0}{
    \begin{tabular}{lcclc}
    \toprule
    Model & Param.(M) & GFLOPs & Top-1 & Reference\\
    \midrule
    DeiT-T~\cite{touvron2021deit} & ~~5.7   & ~~1.3   & 72.2 & ICML21\\
    TnT-Ti~\cite{han2021tnt} & ~~6.1 &    ~~1.4       & 73.9 & NeurIPS21 \\
    ViL-Ti-RPB~\cite{zhang2021vil} & ~~6.7 & ~~1.3    & 76.7 & ICCV21 \\
    XCiT-T12/16~\cite{el2021xcit} & ~~7.0   & ~~1.2   & 77.1 & NeurIPS21 \\
    ViTAE-6M~\cite{xu2021vitae}   & ~~6.5   & ~~2.0   & 77.9 & NeurIPS21 \\
    \rowcolor{whitesmoke}CoaT-Lite T~\cite{xu2021coat} & ~~5.7   & ~~1.6   & 76.6 & ICCV21 \\
    \rowcolor{powderblue}\textbf{MPViT-T} & ~~5.8   & ~~1.6   & \textbf{78.2~(+1.6)} & \\
    \midrule
    ResNet-18~\cite{he2016resnet} & 11.7  & ~~1.8   & 69.8 &CVPR16 \\
    PVT-T~\cite{wang2021pvt} & 13.2  & ~~1.9   & 75.1 & ICCV21 \\
    XCiT-T24/16~\cite{el2021xcit} & 12.0  & ~~2.3   & 79.4 & NeurIPS21 \\
    CoaT Mi~\cite{xu2021coat} & 10.0  & ~~6.8   & 80.8 & ICCV21 \\
    \rowcolor{whitesmoke}CoaT-Lite Mi~\cite{xu2021coat} & 11.0  & ~~2.0   & 78.9 & ICCV21 \\
    \rowcolor{powderblue}\textbf{MPViT-XS} & 10.5  & ~~2.9   & \textbf{80.9~(+2.0)}& \\
    \midrule
    ResNet-50~\cite{he2016resnet} & 25.6  & ~~4.1   & 76.1 &CVPR16 \\
    PVT-S~\cite{wang2021pvt} & 24.5  & ~~3.8   & 79.8 & ICCV21 \\
    DeiT-S/16~\cite{touvron2021deit} & 22.1  & ~~4.6   & 79.9 & ICML21 \\
    Swin-T~\cite{liu2021swin} & 29.0  & ~~4.5   & 81.3 & ICCV21 \\
    Twins-SVT-S~\cite{chu2021Twins} &24.0 & 2.8 & 81.3 & NeurIPS21 \\
    TnT-S~\cite{han2021tnt} & 23.8 &    ~~5.2       & 81.5 & NeurIPS21 \\
    CvT-13~\cite{wu2021cvt} & 20.0  & ~~4.5   & 81.6 & ICCV21 \\
    XCiT-S12/16~\cite{el2021xcit} & 26.0  & ~~4.8   & 82.0 & NeurIPS21 \\
    ViTAE-S~\cite{xu2021vitae} & 23.6 & ~~5.6       & 82.0 & NeurIPS21 \\
    GG-T~\cite{yu2021gg}       & 28.0 & ~~4.5       & 82.0 & NeurIPS21 \\
    CoaT S~\cite{xu2021coat} & 22.0  & 12.6  & 82.1 & ICCV21 \\
    Focal-T~\cite{yang2021focal} & 29.1  & ~~4.9   & 82.2 & NeurIPS21 \\
    CrossViT-15~\cite{chen2021crossvit} & 28.2 & ~~6.1 & 82.3 & ICCV21 \\
    ViL-S-RPB~\cite{zhang2021vil} & 24.6 & ~~4.9    & 82.4 & ICCV21 \\
    CvT-21~\cite{wu2021cvt} & 32.0  & ~~7.1   & 82.5 & ICCV21 \\
    CrossViT-18~\cite{chen2021crossvit} & 43.3 & ~~9.5 & 82.8 & ICCV21 \\
    HRFormer-B~\cite{yuan2021hrformer} & 50.3 & 13.7 & 82.8 & NeurIPS21 \\
    \rowcolor{whitesmoke}CoaT-Lite S~\cite{xu2021coat} & 20.0  & ~~4.0   & 81.9 & ICCV21 \\
    \rowcolor{powderblue}\textbf{MPViT-S} & 22.8  & ~~4.7   & \textbf{83.0~(+1.1)}& \\
    \midrule
    ResNeXt-101~\cite{xie2017resnext} & 83.5  & 15.6  & 79.6 &CVPR17 \\
    PVT-L~\cite{wang2021pvt} & 61.4  & ~~9.8   & 81.7 & ICCV21 \\
    DeiT-B/16~\cite{touvron2021deit} & 86.6  & 17.6  & 81.8 & ICML21 \\
    XCiT-M24/16~\cite{el2021xcit} & 84.0  & 16.2  & 82.7 & NeurIPS21 \\
    Twins-SVT-B~\cite{chu2021Twins} & 56.0 & ~~8.3 & 83.1 & NeurIPS21 \\ 
    Swin-S~\cite{liu2021swin} & 49.6  & ~~8.7  & 83.1 & ICCV21 \\
    Twins-SVT-L~\cite{chu2021Twins} & 99.2 & 14.8 & 83.3 & NeurIPS21 \\ 
    Swin-B~\cite{liu2021swin} & 88.0  & 15.4  & 83.3 & ICCV21 \\
    XCiT-S12/8~\cite{el2021xcit} & 26.0  & 18.9  & 83.4 & NeurIPS21 \\
    Focal-S~\cite{yang2021focal} & 51.1  & ~9.1  & 83.5  & NeurIPS21 \\
    XCiT-M24/8~\cite{el2021xcit} & 84.0  & 63.9  & 83.7 & NeurIPS21 \\
    Focal-B~\cite{yang2021focal} & 89.8  & 16.0  & 83.8  & NeurIPS21 \\
    XCiT-S24/8~\cite{el2021xcit} & 48.0  & 36.0  & 83.9 & NeurIPS21 \\
    \rowcolor{powderblue}\textbf{MPViT-B} & 74.8  & 16.4  & \textbf{84.3}&  \\
    \bottomrule
    \end{tabular}%
    }
    \end{adjustbox}
 \caption{\textbf{Full comparison on ImageNet-1K classification.} These models are trained with $224\times224$ resolution. For fair comparison, we do not include models that are distilled~\cite{touvron2021deit} or use $384\times384$ resolution. Note that CoaT-Lite~\cite{xu2021coat} models are our single-path baselines.}
  \label{tab:cls_supp}%
    \vspace{-0.5cm}
\end{table}%

\begin{figure*}
\begin{center}
\scalebox{1.0}{
\includegraphics[width=0.92\textwidth]{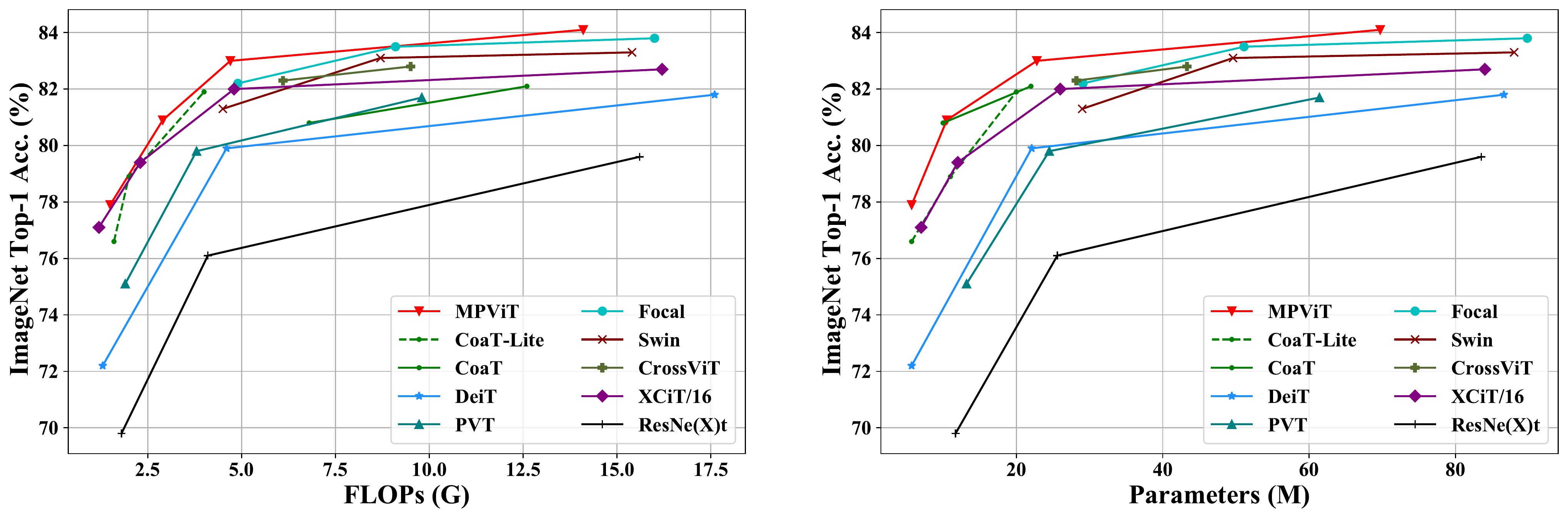}
}
\end{center}
\caption{ \textbf{Performance comparisons with respect to FLOPs and model parameters on ImageNet-1K classification.} 
These models are trained with $224\times224$ single-crop. For fair comparison, we do not include models that are distilled~\cite{touvron2021deit} or use $384\times384$ resolution.
}
\label{fig:cls_plot}
\end{figure*}

\paragraph{Object detection and Instance segmentation}
For fair comparison, we follow the training recipe as in CoaT~\cite{xu2021coat} and Swin Transformer~\cite{liu2021swin} for RetinaNet~\cite{lin2017retinanet} and Mask R-CNN~\cite{he2017mask}.
Specifically, we train all models for $3\times$ schedule~(36 epochs)~\cite{he2017mask,wu2019detectron2} with multi-scale inputs~(MS)~\cite{carion2020detr,sun2021sparsercnn} which resizes the input such that the shorter side is between 480 and 800 while the longer side is at most 1333).
We use the AdamW~\cite{loshchilov2017adamw} optimizer, a weight decay of 0.05, a batch size of 16, and an initial learning rate of 0.0001 which is decayed by $10\times$ at epochs 27 and 33.
We set stochastic depth drop rates~\cite{huang2016dpr} to 0.1, 0.1, 0.2, and 0.4 for Tiny, XSmall, Small, and Base, respectively.
We implement all models based on the \texttt{detectron2} library~\cite{wu2019detectron2}.

\paragraph{Semantic segmentation}
Following the same training recipe as in Swin Transformer~\cite{liu2021swin} and XCiT~\cite{el2021xcit}, we deploy UperNet~\cite{xiao2018upernet} with the AdamW~\cite{loshchilov2017adamw} optimizer, a weight decay of 0.01, an initial learning rate of $6\times10^{-5}$ which is scaled using a linear learning rate decay, and linear warmup of 1,500 iterations.
We train models for 160K iterations with a batch size of 16 and an input size of $512\times512$.
We use the same data augmentations as~\cite{liu2021swin, mmseg2020}, utilizing random horizontal flipping, a random re-scaling ratio in the range~[0,5, 2.0] and random photometric distortions.
We set stochastic depth drop rates~\cite{huang2016dpr} to 0.2 and 0.4 for Small and Base, respectively.
We implement all models based on the \texttt{mmseg} library~\cite{mmseg2020}.

\begin{table}[t]
  \centering
  \begin{adjustbox}{width=\columnwidth, center}
    \begin{tabular}{lcccccc}
    \toprule
    Backbone & $AP$ & $AP_{50}$ & $AP_{75}$ & $AP_S$ & $AP_M$ & $AP_L$ \\
    \midrule
    ResNet-50~\cite{he2016resnet} & 44.5  & 63.7  & 48.7  & 26.8  & 47.6  & 59.6 \\
    CoaT-Lite small~\cite{xu2021coat} & 47.0  & 66.5  & 51.2  & 28.8  & 50.3  & 63.3 \\
    CoaT Small~\cite{xu2021coat} & 48.4  & 68.5  & 52.4  & 30.2  & 51.8  & 63.8 \\
    \rowcolor{powderblue}\textbf{MPViT-Small} & \textbf{49.0}  & \textbf{68.7}  & \textbf{53.7}  & \textbf{31.7}  & \textbf{52.4}  & \textbf{64.5} \\
    \bottomrule
    \end{tabular}%
    \end{adjustbox}
    \caption{\textbf{COCO Object Detection results on Deformable DETR}~\cite{zhu2020deform-detr}. These all models are trained using the same codebase.}
    
  \label{tab:dd}%
\end{table}%

\subsection{More Experimental Analysis}
\label{sec:ex}

\paragraph{ImageNet classification}
We provide a full summary of comparisons on ImageNet-1K classification in \Table{cls_supp} by adding more recent Vision Transformers including ViL~\cite{zhang2021vil}, TnT~\cite{han2021tnt}, ViTAE~\cite{xu2021vitae}, HRFormer~\cite{yuan2021hrformer}, and Twins~\cite{chu2021Twins}.
We can observe that MPViTs consistently achieve state-the-art performance compared to SOTA models with similar model capacity.
Notably, the smaller MPViT variants often outperform their larger baseline counterparts even when the baselines use significantly more parameters, as shown in \Table{cls_supp} and \Fig{cls_plot}~(right).
Furthermore, \Fig{cls_plot} demonstrates that MPViT is a more \textit{efficient} and \textit{effective} Vision Transformer architecture in terms of computation and model parameters. 

\paragraph{Deformable-DETR}
Additionally, we compare our MPViT-Small with baselines, CoaT-Lite Small~\cite{xu2021coat} and CoaT Small~\cite{xu2021coat}, on the Deformable DETR~(DD)~\cite{zhu2020deform-detr}.
For fair comparison, we train MPViT for 50 epochs with the same training recipe\footnote{\url{https://github.com/mlpc-ucsd/CoaT/tree/main/tasks/Deformable-DETR}} as in CoaT~\cite{xu2021coat}.
We use the AdamW~\cite{loshchilov2017adamw} optimizer with a batch size of 16, a weight decay of $10^{-4}$, and an initial learning rate of $2\times10^{-4}$, which is decayed by a factor of 10 at 40 epoch. 
\Tab{dd} shows results comparing with CoaT-Lite Small and CoaT Small.
MPViT-Small improves over both CoaT-Lite Small and CoaT Small.
Notably, MPViT achieves a larger gain in small object AP~(1.5\% $AP_{S}$) as compared to others~(\ie, $AP_{M}$ or $AP_{L}$).

\begin{table*}[htbp]
  \centering
      \begin{adjustbox}{max width=\textwidth}\
\begin{tabular}{lll|cccccc|cccccc}
    \toprule
    \multirow{2}[2]{*}{Backbone} & \multirow{2}[2]{*}{Params. (M)} & \multirow{2}[2]{*}{GFLOPs}  & \multicolumn{6}{c|}{Mask R-CNN $1\times$} & \multicolumn{6}{c}{RetinaNet $1\times$} \\
          &       &             & $AP^b$ & $AP^b_{50}$ & $AP^b_{75}$ & $AP^m$ & $AP^m_{50}$ & $AP^m_{75}$ & $AP^b$ & $AP^b_{50}$ & $AP^b_{75}$ & $AP^b_S$ & $AP^b_M$ & $AP^b_L$ \\
    \midrule
    PVTv2-B0~\cite{wang2021pvtv2} & ~~23 (13)    & ~~195 (177)    & 38.2  & 60.5  &40.7 & 36.2  & 57.8  & 38.6  & 37.2     & 57.2     & 39.5     & 23.1     & 40.4     & 49.7 \\
    \rowcolor{powderblue} \textbf{MPViT-T} & ~~28 (17) & ~~216 (196)  & \textbf{42.2}  & 64.2  & 45.8  & \textbf{39.0}  & 61.4  & 41.8  & \textbf{41.8}  & 62.7  & 44.6  & 27.2  & 45.1  & 54.2  \\
    \midrule
    PVT-T~\cite{wang2021pvt} & ~~33 (23) & ~~240 (221)   & 39.8  & 62.2  & 43.0  & 37.4  & 59.3  & 39.9  & 39.4  & 59.8  & 42.0  & 25.5  & 42.0  & 52.1  \\
    PVTv2-B1~\cite{wang2021pvtv2} & ~~33 (23) & ~~243 (225)   & 41.8  & 54.3  & 45.9  & 38.8  & 61.2  & 41.6  & 41.2  & 61.9  & 43.9  & 25.4  & 44.5  & 54.3  \\
    \rowcolor{powderblue} \textbf{MPViT-XS} & ~~30 (20) & ~~231 (211)   & \textbf{44.2}  & 66.7  & 48.4  & \textbf{40.4}  & 63.4  & 43.4  & \textbf{43.8}  & 65.0  & 47.1  & 28.1  & 47.6  & 56.5  \\
    \midrule
    ResNet-50~\cite{he2016resnet} & ~~44 (38) & ~~260 (239) & 38.0  & 58.6  & 41.4  & 34.4  & 55.1  & 36.7  & 36.3  & 55.3  & 38.6  & 19.3  & 40.4  & 48.8  \\
    PVT-S~\cite{wang2021pvt} & ~~44 (34) & ~~305 (226)   & 43.0  & 65.3  & 46.9  & 39.9  & 62.5  & 42.8  & 42.2  & 62.7  & 45.0  & 26.2  & 45.2  & 57.2  \\
    PVTv2-B2~\cite{wang2021pvtv2} & ~~45 (35) & ~~309 (290)   & 45.3  & 67.1  & 49.6  & 41.2  & 64.2  & 44.4  & 44.6  & 65.6  & 47.6  & 27.4  & 48.8  & 58.6  \\
    Swin-T~\cite{liu2021swin} & ~~48 (39) & ~~267 (245)   & 43.7  & 66.6  & 47.7  & 39.8  & 63.3  & 42.7  & 42.0  & 63.0  & 44.7  & 26.6  & 45.8  & 55.7  \\
    Focal-T~\cite{yang2021focal} & ~~49 (39) & ~~291 (265)      & 44.8  & 67.7  & 49.2  & 41.0  & 64.7  & 44.2  & 43.7  & 65.2  & 46.7  & 28.6  & 47.4  & 56.9  \\
    \rowcolor{powderblue} \textbf{MPViT-S} & ~~43 (32) & ~~268 (248)    & \textbf{46.4}  &68.6  & 51.2  & \textbf{42.4}  & 65.6  & 45.7  & \textbf{45.7}  & 57.3  & 48.8  & 28.7  & 49.7  & 59.2  \\
    \midrule
    ResNeXt101-64x4d~\cite{xie2017resnext} & ~102 (96) & ~~493 (473)   & 42.8  & 63.8  & 47.3  & 38.4  & 60.6  & 41.3  & 41.0  & 60.9  & 44.0  & 23.9  & 45.2  & 54.0  \\
    PVT-M~\cite{wang2021pvt} & ~~64 (54) & ~~392 (283)    & 42.0  & 64.4  & 45.6  & 39.0  & 61.6  & 42.1  & 41.9  & 63.1  & 44.3  & 25.0  & 44.9  & 57.6  \\
    PVT-L~\cite{wang2021pvt} & ~~81 (71) & ~~494 (345)    & 42.9  & 65.0  & 46.6  & 39.5  & 61.9  & 42.5  & 42.6  & 63.7  & 45.4  & 25.8  & 46.0  & 58.4  \\
    PVTv2-B5~\cite{wang2021pvtv2} & ~101 (91) & ~~557 (538)  & 47.4  & 68.6  & 51.9  & 42.5  & 65.7  & 46.0  & 46.2  & 67.1  & 49.5  & 28.5  & 50.0  & 62.5  \\
    Swin-S~\cite{liu2021swin} & ~~69 (60) & ~~359 (335)   & 46.5  & 68.7  & 51.3  & 42.1  & 65.8  & 45.2  & 45.0  & 66.2  & 48.3  & 27.9  & 48.8  & 59.5  \\
    Swin-B~\cite{liu2021swin} & ~107 (98) & ~~496 (477)      & 46.9  & 69.2  & 51.6  & 42.3  & 66.0  & 45.5  & 45.0  & 66.4  & 48.3  & 28.4  & 49.1  & 60.6  \\
    Focal-S~\cite{yang2021focal} & ~~71 (62) & ~~401 (367)   & 47.4  & 69.8  & 51.9  & 42.8  & 66.6  & 46.1  & 45.6  & 67.0  & 48.7  & 29.5  & 49.5  & 60.3  \\
    Focal-B~\cite{yang2021focal} & ~~110 (101) & ~~533 (514)      & 47.8  & 70.2  & 52.5  & 43.2  & 67.3  & 46.5  & 46.3  & 68.0  & 49.8  & 31.7  & 50.4  & 60.8  \\
    \rowcolor{powderblue} \textbf{MPViT-B} & ~~95 (85) & ~~503 (482)  & \textbf{48.2}      & 70.0      & 52.9       &\textbf{43.5}       &67.1     &46.8      &\textbf{47.0}       &68.4     &50.8       &29.4       &51.3       &61.5  \\
    \bottomrule
    \end{tabular}%
    \end{adjustbox}
    \caption{\textbf{COCO detection and instance segmentation} with RetinaNet~\cite{lin2017retinanet} and Mask R-CNN~\cite{he2017mask}. Models are trained for $1\times$ schedule~\cite{wu2019detectron2} with single-scale training inputs.
    All backbones are pretrained on ImageNet-1K. 
    We omit models pretrained on larger-datasets~(\eg, ImageNet-21K). 
    The GFLOPs are measured at resolution $800\times1280$.
    Mask R-CNN's parameters/FLOPs  are followed by RetinaNet in parentheses.}
  \label{tab:det_supp}%
\end{table*}%

\paragraph{COCO with $1\times$ schedule.}
In addition to the $3\times$ schedule + multi-scale~(MS) setting, we also evaluate MPViT on RetinaNet~\cite{lin2017retinanet} and Mask R-CNN~\cite{he2017mask} with $1\times$ schedule~(12 epochs)~\cite{wu2019detectron2} using single-scale inputs.
\Tab{det_supp} shows result comparisons with state-of-the-art methods.
In the results of $3\times$ schedule + multi-scale~(MS), we can also observe that MPViTs consistently outperform on both RetinaNet and Mask R-CNN. 
We note that MPViTs surpass the most recent improved PVTv2~\cite{wang2021pvtv2} models.

\subsection{More Qualitative Results}
\label{sec:vis_supp}
\begin{figure*}
\begin{center}
\scalebox{1.0}{
\includegraphics[width=\textwidth]{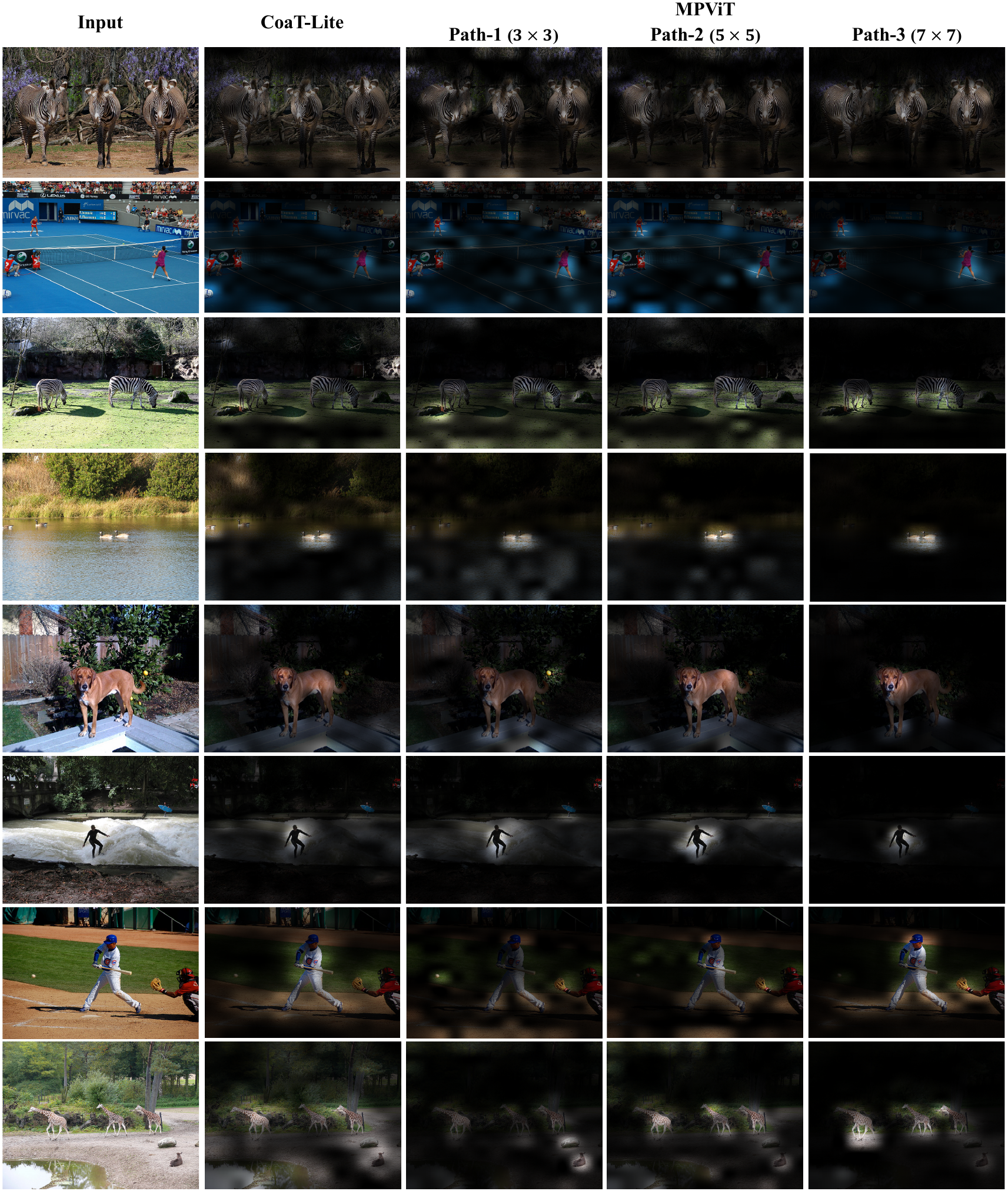}
}
\end{center}
\caption{\textbf{Additional Attention Maps} generated by CoaT-Lite~\cite{xu2021coat} and our MPViT.
MPViT has a triple-path structure with patches of various sizes~(\eg, $3\times3, 5\times5, 7\times7$), leading to fine and coarse features.
}
\label{fig:att_supp}
\end{figure*}

\paragraph{Visualization of Attention Maps}
As shown in Eq.(\ref{eq:factatt}), the factorized self-attention in~\cite{xu2021coat} first extracts channel-wise attention softmax$(K)$ by applying a softmax over spatial dimensions~(x,~y).
Then, softmax$(K)^T V$ is computed as below:
\begin{equation}
\begin{split}
    & (\text{softmax}(K)^TV)(c_i, c_j) \\
    & = \sum _ {(x,y)} \text{softmax}(K)(x,y,c_i) V(x, y, c_j),
    \label{eq:ktv}
\end{split}
\end{equation}
where $x$ and $y$ are position of tokens. 
$c_i$ and $c_j$ indicate channel indices of $K$ and $V$, respectively.
It can be interpreted as multiplying $V$ by the channel-wise spatial attention in a pixel-wise manner followed by the sum over spatial dimension.
In other words, softmax$(K)^T V$ represents the weighted sum of V where the weight of each position $(x,y)$ is the channel-wise spatial attention.
Therefore, to obtain the importance of each position, we employ the mean of softmax$(K)$ over the channel dimension, resulting in spatial attention.
Then, the spatial attention is overlaid to the original input image for better visualization, as shown in Fig.~\ref{fig:att_supp}.
In detail, we resize the spatial attention to the size of the original image, normalize the value to [0,1], and then multiply the attention map by the image.
 
To validate the effectiveness of our attention map qualitatively, we compare attention maps of MPViT and CoaT-Lite~\cite{xu2021coat} in Fig.~\ref{fig:att_supp}.
We compare the attention maps of each method generated from the 4th stage in the same way.
For a fair comparison, we pick the best qualitative attention map of each method since both CoaT-Lite and MPViT have eight heads for each layer.
Furthermore, we visualize attention maps extracted from all three paths of MPViT to observe the individual effects of each path.

As mentioned in Section~\ref{sec:vis}, the three paths of MPViT can capture objects of varying sizes due to the multi-scale embedding of MPViT as the similar effect of multiple receptive fields.
In other words, path-1 concentrates on small objects or textures while path-3 focuses on large objects or high-level semantic concepts.
We support this intuition by observing more examples shown in Fig.~\ref{fig:att_supp}.
Attention maps of path-1 (3rd column) capture small objects such as small ducks (4th row), an orange (5th row), a small ball (6th row), and an antelope (8th row).
In addition, since path-1 also captures textures due to a smaller receptive field, a relatively low level of attention is present in the background.
In contrast, we can observe different behavior for path-3, which can be seen in the rightmost column.
Path-3 accentuates large objects while suppressing the background and smaller objects. 
For example, the ducks (4th row), orange (5th row), and ball (8th row) are masked out in the rightmost column since path-3 concentrates on larger objects.
The attention maps of path-2 (4th column) showcase the changing behavior between paths-1 and 3 since the scale of path-2 is in-between the scales of paths-1 and 3, and accordingly, the attention maps also begin to transition from smaller to larger objects.
In other words, although the attention map of path-2 attends similar regions as path-1, it is also more likely to emphasize larger objects, as path-3 does.
For example, in the last row, path-2 attends to similar regions as path-1 while emphasizing the large giraffes more than path-1.
Therefore, although the three paths independently deal with different scales, they act in a complementary manner, which is beneficial for dense prediction tasks.

Since Coat-Lite has a single-path architecture, the singular path needs to deal with objects of varying sizes. 
Therefore, attention maps from CoaT-Lite~(2nd column) simultaneously attend to large and small objects, as shown in the 4th row. 
However, it is difficult to capture all objects with a single path, as CoaT-Lite misses the orange (5th row) and ball (7th row).
In addition, Coat-Lite cannot capture object boundaries as precisely as path-3 of MPViT since path-3 need not attend to small objects or textures.
As a result, MPViT shows superior results compared to Coat-Lite on classification, detection, and segmentation tasks.

\begin{figure*}
\begin{center}
\scalebox{1.0}{
\includegraphics[width=\textwidth]{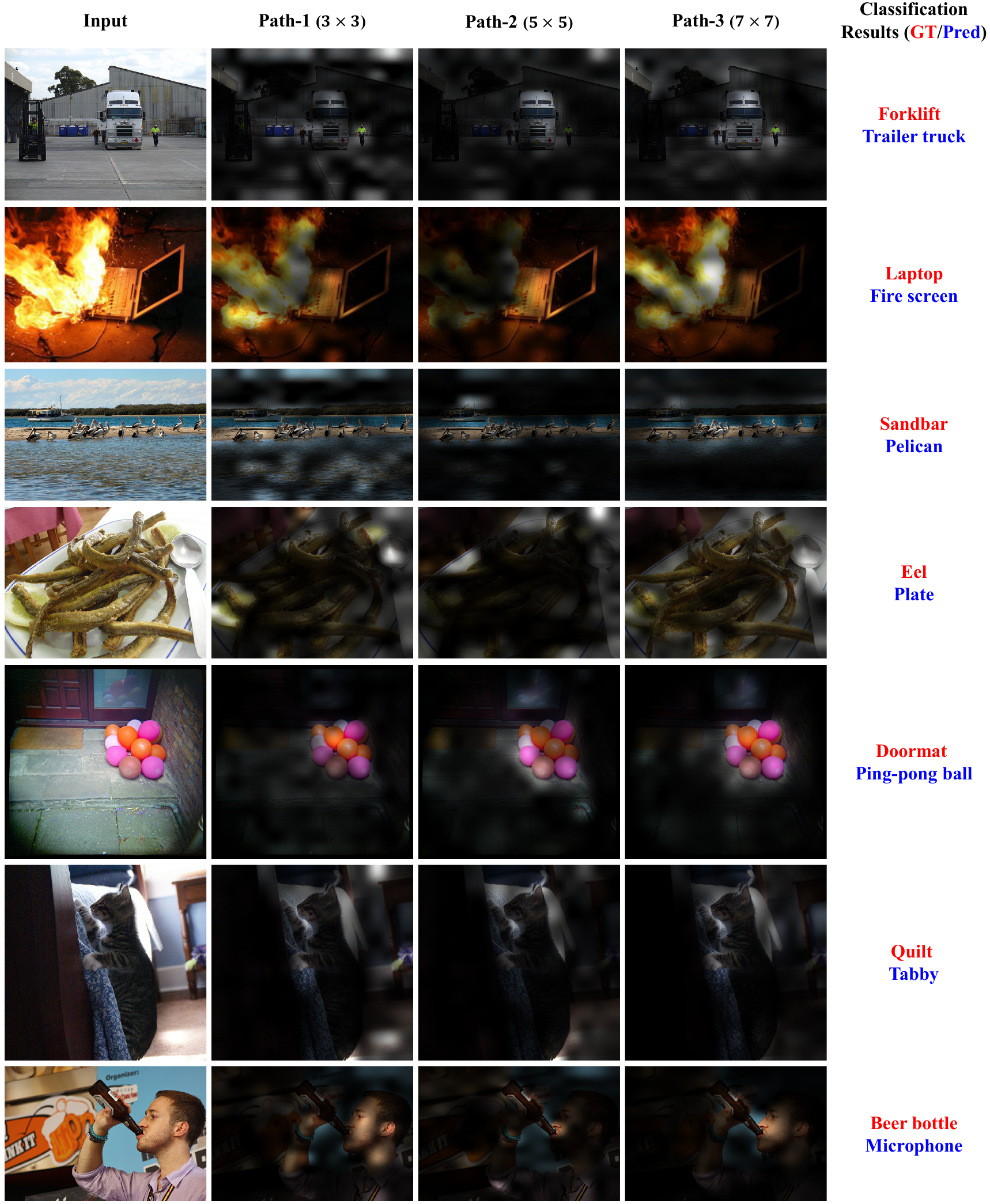}
}
\end{center}
\caption{\textbf{Attention Maps of failure cases on ImageNet \textit{validation} images.} 
The input image and corresponding attention maps from each path are illustrated. In the rightmost column, we show the ground truth labels and predicted labels colored with red and blue, respectively.
}
\label{fig:failure}
\end{figure*}

\paragraph{Failure case}
In order to verify the effects of attention from a different perspective, we further analyze failure cases on the ImageNet \textit{validation} images. 
We show attention maps of each path corresponding to the input image along with the ground truth and the predicted labels of MPViT in Fig.~\ref{fig:failure}.
For example, in the first row, the ground truth of the input image is a forklift, while the predicted label is a trailer truck. 
Although the attention map from path-1 places light emphasis on the forklift, the 
attention maps from all paths commonly accentuate the trailer truck rather than the forklift, which leads to classifying the image as a trailer truck and not a forklift.
Other classification results in Fig.~\ref{fig:failure} fail in similar circumstances, except for the last row.
In the last row, MPViTs attention maps correctly capture the beer bottle. 
However, the attention maps also attend to the face near the bottle.
Therefore, the bottle is misunderstood as a microphone since the image of ``drinking a bottle of beer" and ``using a microphone" are semantically similar.
From the above, we can observe that the attention maps and the predicted results are highly correlated.

{\small
\bibliographystyle{ieee_fullname}
\bibliography{main}

\begin{thebibliography}{10}\itemsep=-1pt

\bibitem{arnab2021vivit}
Anurag Arnab, Mostafa Dehghani, Georg Heigold, Chen Sun, Mario Lu{\v{c}}i{\'c},
  and Cordelia Schmid.
\newblock Vivit: A video vision transformer.
\newblock In {\em ICCV}, 2021.

\bibitem{baker2018shape}
Nicholas Baker, Hongjing Lu, Gennady Erlikhman, and Philip~J Kellman.
\newblock Deep convolutional networks do not classify based on global object
  shape.
\newblock {\em PLoS computational biology}, 14(12):e1006613, 2018.

\bibitem{bertasius2021timesformer}
Gedas Bertasius, Heng Wang, and Lorenzo Torresani.
\newblock Is space-time attention all you need for video understanding?
\newblock In {\em ICML}, 2021.

\bibitem{brown2020gpt3}
Tom~B Brown, Benjamin Mann, Nick Ryder, Melanie Subbiah, Jared Kaplan, Prafulla
  Dhariwal, Arvind Neelakantan, Pranav Shyam, Girish Sastry, Amanda Askell,
  et~al.
\newblock Language models are few-shot learners.
\newblock {\em arXiv preprint arXiv:2005.14165}, 2020.

\bibitem{carion2020detr}
Nicolas Carion, Francisco Massa, Gabriel Synnaeve, Nicolas Usunier, Alexander
  Kirillov, and Sergey Zagoruyko.
\newblock End-to-end object detection with transformers.
\newblock In {\em ECCV}, 2020.

\bibitem{chen2021crossvit}
Chun-Fu Chen, Quanfu Fan, and Rameswar Panda.
\newblock Crossvit: Cross-attention multi-scale vision transformer for image
  classification.
\newblock In {\em ICCV}, 2021.

\bibitem{chen2021xtracking}
Xin Chen, Bin Yan, Jiawen Zhu, Dong Wang, Xiaoyun Yang, and Huchuan Lu.
\newblock Transformer tracking.
\newblock In {\em CVPR}, 2021.

\bibitem{Chen_2021visformer}
Zhengsu Chen, Lingxi Xie, Jianwei Niu, Xuefeng Liu, Longhui Wei, and Qi Tian.
\newblock Visformer: The vision-friendly transformer.
\newblock In {\em ICCV}, 2021.

\bibitem{chollet2017xception}
Fran{\c{c}}ois Chollet.
\newblock Xception: Deep learning with depthwise separable convolutions.
\newblock In {\em CVPR}, 2017.

\bibitem{chu2021Twins}
Xiangxiang Chu, Zhi Tian, Yuqing Wang, Bo Zhang, Haibing Ren, Xiaolin Wei,
  Huaxia Xia, and Chunhua Shen.
\newblock Twins: Revisiting the design of spatial attention in vision
  transformers.
\newblock In {\em NeurIPS}, 2021.

\bibitem{mmseg2020}
MMSegmentation Contributors.
\newblock {MMSegmentation}: Openmmlab semantic segmentation toolbox and
  benchmark.
\newblock \url{https://github.com/open-mmlab/mmsegmentation}, 2020.

\bibitem{dai2021up}
Zhigang Dai, Bolun Cai, Yugeng Lin, and Junying Chen.
\newblock Up-detr: Unsupervised pre-training for object detection with
  transformers.
\newblock In {\em CVPR}, 2021.

\bibitem{deng2009imagenet}
Jia Deng, Wei Dong, Richard Socher, Li-Jia Li, Kai Li, and Li Fei-Fei.
\newblock Imagenet: A large-scale hierarchical image database.
\newblock In {\em CVPR}, 2009.

\bibitem{devlin2019bert}
Jacob Devlin, Ming-Wei Chang, Kenton Lee, and Kristina Toutanova.
\newblock Bert: Pre-training of deep bidirectional transformers for language
  understanding.
\newblock In {\em NAACL-HLT (1)}, 2019.

\bibitem{dollar2021fast}
Piotr Doll{\'a}r, Mannat Singh, and Ross Girshick.
\newblock Fast and accurate model scaling.
\newblock In {\em CVPR}, 2021.

\bibitem{dosovitskiy2021vit}
Alexey Dosovitskiy, Lucas Beyer, Alexander Kolesnikov, Dirk Weissenborn,
  Xiaohua Zhai, Thomas Unterthiner, Mostafa Dehghani, Matthias Minderer, Georg
  Heigold, Sylvain Gelly, et~al.
\newblock An image is worth 16x16 words: Transformers for image recognition at
  scale.
\newblock In {\em ICLR}, 2021.

\bibitem{el2021xcit}
Alaaeldin El-Nouby, Hugo Touvron, Mathilde Caron, Piotr Bojanowski, Matthijs
  Douze, Armand Joulin, Ivan Laptev, Natalia Neverova, Gabriel Synnaeve, Jakob
  Verbeek, et~al.
\newblock Xcit: Cross-covariance image transformers.
\newblock In {\em NeurIPS}, 2021.

\bibitem{Fan_2021mvit}
Haoqi Fan, Bo Xiong, Karttikeya Mangalam, Yanghao Li, Zhicheng Yan, Jitendra
  Malik, and Christoph Feichtenhofer.
\newblock Multiscale vision transformers.
\newblock In {\em ICCV}, 2021.

\bibitem{gao2019res2net}
Shanghua Gao, Ming-Ming Cheng, Kai Zhao, Xin-Yu Zhang, Ming-Hsuan Yang, and
  Philip~HS Torr.
\newblock Res2net: A new multi-scale backbone architecture.
\newblock {\em TPAMI}, 2019.

\bibitem{graham2021levit}
Ben Graham, Alaaeldin El-Nouby, Hugo Touvron, Pierre Stock, Armand Joulin,
  Herv{\'e} J{\'e}gou, and Matthijs Douze.
\newblock Levit: a vision transformer in convnet's clothing for faster
  inference.
\newblock In {\em ICCV}, 2021.

\bibitem{han2021tnt}
Kai Han, An Xiao, Enhua Wu, Jianyuan Guo, Chunjing Xu, and Yunhe Wang.
\newblock Transformer in transformer.
\newblock In {\em NeurIPS}, 2021.

\bibitem{he2017mask}
Kaiming He, Georgia Gkioxari, Piotr Doll{\'a}r, and Ross Girshick.
\newblock Mask r-cnn.
\newblock In {\em CVPR}, 2017.

\bibitem{he2016resnet}
Kaiming He, Xiangyu Zhang, Shaoqing Ren, and Jian Sun.
\newblock Deep residual learning for image recognition.
\newblock In {\em CVPR}, 2016.

\bibitem{hoffer2020augment}
Elad Hoffer, Tal Ben-Nun, Itay Hubara, Niv Giladi, Torsten Hoefler, and Daniel
  Soudry.
\newblock Augment your batch: Improving generalization through instance
  repetition.
\newblock In {\em CVPR}, 2020.

\bibitem{howard2019mobilenetv3}
Andrew Howard, Mark Sandler, Grace Chu, Liang-Chieh Chen, Bo Chen, Mingxing
  Tan, Weijun Wang, Yukun Zhu, Ruoming Pang, Vijay Vasudevan, et~al.
\newblock Searching for mobilenetv3.
\newblock In {\em CVPR}, 2019.

\bibitem{howard2017mobilenet}
Andrew~G Howard, Menglong Zhu, Bo Chen, Dmitry Kalenichenko, Weijun Wang,
  Tobias Weyand, Marco Andreetto, and Hartwig Adam.
\newblock Mobilenets: Efficient convolutional neural networks for mobile vision
  applications.
\newblock {\em arXiv preprint arXiv:1704.04861}, 2017.

\bibitem{huang2016dpr}
Gao Huang, Yu Sun, Zhuang Liu, Daniel Sedra, and Kilian~Q Weinberger.
\newblock Deep networks with stochastic depth.
\newblock In {\em ECCV}, 2016.

\bibitem{inoue2018mixup}
Hiroshi Inoue.
\newblock Data augmentation by pairing samples for images classification.
\newblock {\em arXiv preprint arXiv:1801.02929}, 2018.

\bibitem{ioffe2015bn}
Sergey Ioffe and Christian Szegedy.
\newblock Batch normalization: Accelerating deep network training by reducing
  internal covariate shift.
\newblock In {\em ICML}, 2015.

\bibitem{islam2020much}
Md~Amirul Islam, Sen Jia, and Neil~DB Bruce.
\newblock How much position information do convolutional neural networks
  encode?
\newblock {\em arXiv preprint arXiv:2001.08248}, 2020.

\bibitem{kayhan2020translation}
Osman~Semih Kayhan and Jan C~van Gemert.
\newblock On translation invariance in cnns: Convolutional layers can exploit
  absolute spatial location.
\newblock In {\em CVPR}, 2020.

\bibitem{lee2019vovnet}
Youngwan Lee, Joong-won Hwang, Sangrok Lee, Yuseok Bae, and Jongyoul Park.
\newblock An energy and gpu-computation efficient backbone network for
  real-time object detection.
\newblock In {\em CVPRW}, 2019.

\bibitem{lee2017wri}
Youngwan Lee, Huieun Kim, Eunsoo Park, Xuenan Cui, and Hakil Kim.
\newblock Wide-residual-inception networks for real-time object detection.
\newblock In {\em IEEE Intelligent Vehicles Symposium (IV)}, 2017.

\bibitem{lin2017fpn}
Tsung-Yi Lin, Piotr Doll{\'a}r, Ross Girshick, Kaiming He, Bharath Hariharan,
  and Serge Belongie.
\newblock Feature pyramid networks for object detection.
\newblock In {\em CVPR}, 2017.

\bibitem{lin2017retinanet}
Tsung-Yi Lin, Priya Goyal, Ross Girshick, Kaiming He, and Piotr Doll{\'a}r.
\newblock Focal loss for dense object detection.
\newblock In {\em ICCV}, 2017.

\bibitem{lin2014coco}
Tsung-Yi Lin, Michael Maire, Serge Belongie, James Hays, Pietro Perona, Deva
  Ramanan, Piotr Doll{\'a}r, and C~Lawrence Zitnick.
\newblock Microsoft coco: Common objects in context.
\newblock In {\em ECCV}, 2014.

\bibitem{liu2021swin}
Ze Liu, Yutong Lin, Yue Cao, Han Hu, Yixuan Wei, Zheng Zhang, Stephen Lin, and
  Baining Guo.
\newblock Swin transformer: Hierarchical vision transformer using shifted
  windows.
\newblock In {\em ICCV}, 2021.

\bibitem{loshchilov2017adamw}
Ilya Loshchilov and Frank Hutter.
\newblock Decoupled weight decay regularization.
\newblock {\em arXiv preprint arXiv:1711.05101}, 2017.

\bibitem{lowe1999object}
David~G Lowe.
\newblock Object recognition from local scale-invariant features.
\newblock In {\em ICCV}, 1999.

\bibitem{ma2018shufflenetv2}
Ningning Ma, Xiangyu Zhang, Hai-Tao Zheng, and Jian Sun.
\newblock Shufflenet v2: Practical guidelines for efficient cnn architecture
  design.
\newblock In {\em ECCV}, 2018.

\bibitem{meinhardt2021trackformer}
Tim Meinhardt, Alexander Kirillov, Laura Leal-Taixe, and Christoph
  Feichtenhofer.
\newblock Trackformer: Multi-object tracking with transformers.
\newblock {\em arXiv preprint arXiv:2101.02702}, 2021.

\bibitem{newell2016hourglass}
Alejandro Newell, Kaiyu Yang, and Jia Deng.
\newblock Stacked hourglass networks for human pose estimation.
\newblock In {\em ECCV}, 2016.

\bibitem{radford2018gpt}
Alec Radford, Karthik Narasimhan, Tim Salimans, and Ilya Sutskever.
\newblock Improving language understanding with unsupervised learning.
\newblock {\em Technical report, OpenAI}, 2018.

\bibitem{simonyan2014vgg}
Karen Simonyan and Andrew Zisserman.
\newblock Very deep convolutional networks for large-scale image recognition.
\newblock In {\em ICLR}, 2014.

\bibitem{sun2021sparsercnn}
Peize Sun, Rufeng Zhang, Yi Jiang, Tao Kong, Chenfeng Xu, Wei Zhan, Masayoshi
  Tomizuka, Lei Li, Zehuan Yuan, Changhu Wang, et~al.
\newblock Sparse r-cnn: End-to-end object detection with learnable proposals.
\newblock In {\em CVPR}, 2021.

\bibitem{szegedy2017inception}
Christian Szegedy, Sergey Ioffe, Vincent Vanhoucke, and Alexander~A Alemi.
\newblock Inception-v4, inception-resnet and the impact of residual connections
  on learning.
\newblock In {\em AAAI}, 2017.

\bibitem{szegedy2015googlenet}
Christian Szegedy, Wei Liu, Yangqing Jia, Pierre Sermanet, Scott Reed, Dragomir
  Anguelov, Dumitru Erhan, Vincent Vanhoucke, and Andrew Rabinovich.
\newblock Going deeper with convolutions.
\newblock In {\em CVPR}, 2015.

\bibitem{szegedy2016rethinking}
Christian Szegedy, Vincent Vanhoucke, Sergey Ioffe, Jon Shlens, and Zbigniew
  Wojna.
\newblock Rethinking the inception architecture for computer vision.
\newblock In {\em CVPR}, 2016.

\bibitem{tan2019efficientnet}
Mingxing Tan and Quoc Le.
\newblock Efficientnet: Rethinking model scaling for convolutional neural
  networks.
\newblock In {\em ICML}, 2019.

\bibitem{touvron2021deit}
Hugo Touvron, Matthieu Cord, Matthijs Douze, Francisco Massa, Alexandre
  Sablayrolles, and Herv{\'e} J{\'e}gou.
\newblock Training data-efficient image transformers \& distillation through
  attention.
\newblock In {\em ICML}, 2021.

\bibitem{touvron2021cait}
Hugo Touvron, Matthieu Cord, Alexandre Sablayrolles, Gabriel Synnaeve, and
  Herv{\'e} J{\'e}gou.
\newblock Going deeper with image transformers.
\newblock In {\em ICCV}, 2021.

\bibitem{tuli2021human}
Shikhar Tuli, Ishita Dasgupta, Erin Grant, and Thomas~L Griffiths.
\newblock Are convolutional neural networks or transformers more like human
  vision?
\newblock {\em arXiv preprint arXiv:2105.07197}, 2021.

\bibitem{vaswani2017attention}
Ashish Vaswani, Noam Shazeer, Niki Parmar, Jakob Uszkoreit, Llion Jones,
  Aidan~N Gomez, {\L}ukasz Kaiser, and Illia Polosukhin.
\newblock Attention is all you need.
\newblock In {\em NeurIPS}, 2017.

\bibitem{wang2021maxdeeplab}
Huiyu Wang, Yukun Zhu, Hartwig Adam, Alan Yuille, and Liang-Chieh Chen.
\newblock Max-deeplab: End-to-end panoptic segmentation with mask transformers.
\newblock In {\em CVPR}, 2021.

\bibitem{wang2020hrnet}
Jingdong Wang, Ke Sun, Tianheng Cheng, Borui Jiang, Chaorui Deng, Yang Zhao,
  Dong Liu, Yadong Mu, Mingkui Tan, Xinggang Wang, et~al.
\newblock Deep high-resolution representation learning for visual recognition.
\newblock {\em TPAMI}, 2020.

\bibitem{wang2021tracker}
Ning Wang, Wengang Zhou, Jie Wang, and Houqiang Li.
\newblock Transformer meets tracker: Exploiting temporal context for robust
  visual tracking.
\newblock In {\em ICCV}, 2021.

\bibitem{wang2021pvtv2}
Wenhai Wang, Enze Xie, Xiang Li, Deng-Ping Fan, Kaitao Song, Ding Liang, Tong
  Lu, Ping Luo, and Ling Shao.
\newblock Pvtv2: Improved baselines with pyramid vision transformer.
\newblock {\em arXiv preprint arXiv:2106.13797}, 2021.

\bibitem{wang2021pvt}
Wenhai Wang, Enze Xie, Xiang Li, Deng-Ping Fan, Kaitao Song, Ding Liang, Tong
  Lu, Ping Luo, and Ling Shao.
\newblock Pyramid vision transformer: A versatile backbone for dense prediction
  without convolutions.
\newblock In {\em ICCV}, 2021.

\bibitem{rw2019timm}
Ross Wightman.
\newblock Pytorch image models.
\newblock \url{https://github.com/rwightman/pytorch-image-models}, 2019.

\bibitem{wu2021cvt}
Haiping Wu, Bin Xiao, Noel Codella, Mengchen Liu, Xiyang Dai, Lu Yuan, and Lei
  Zhang.
\newblock Cvt: Introducing convolutions to vision transformers.
\newblock In {\em ICCV}, 2021.

\bibitem{wu2019detectron2}
Yuxin Wu, Alexander Kirillov, Francisco Massa, Wan-Yen Lo, and Ross Girshick.
\newblock Detectron2.
\newblock \url{https://github.com/facebookresearch/detectron2}, 2019.

\bibitem{xiao2018upernet}
Tete Xiao, Yingcheng Liu, Bolei Zhou, Yuning Jiang, and Jian Sun.
\newblock Unified perceptual parsing for scene understanding.
\newblock In {\em ECCV}, 2018.

\bibitem{xie2021segformer}
Enze Xie, Wenhai Wang, Zhiding Yu, Anima Anandkumar, Jose~M Alvarez, and Ping
  Luo.
\newblock Segformer: Simple and efficient design for semantic segmentation with
  transformers.
\newblock In {\em NeurIPS}, 2021.

\bibitem{xie2017resnext}
Saining Xie, Ross Girshick, Piotr Doll{\'a}r, Zhuowen Tu, and Kaiming He.
\newblock Aggregated residual transformations for deep neural networks.
\newblock In {\em CVPR}, 2017.

\bibitem{xu2021coat}
Weijian Xu, Yifan Xu, Tyler Chang, and Zhuowen Tu.
\newblock Co-scale conv-attentional image transformers.
\newblock In {\em ICCV}, 2021.

\bibitem{xu2021vitae}
Yufei Xu, Qiming Zhang, Jing Zhang, and Dacheng Tao.
\newblock Vitae: Vision transformer advanced by exploring intrinsic inductive
  bias.
\newblock In {\em NeurIPS}, 2021.

\bibitem{yang2021focal}
Jianwei Yang, Chunyuan Li, Pengchuan Zhang, Xiyang Dai, Bin Xiao, Lu Yuan, and
  Jianfeng Gao.
\newblock Focal self-attention for local-global interactions in vision
  transformers.
\newblock In {\em NeurIPS}, 2021.

\bibitem{yu2021gg}
Qihang Yu, Yingda Xia, Yutong Bai, Yongyi Lu, Alan Yuille, and Wei Shen.
\newblock Glance-and-gaze vision transformer.
\newblock In {\em NeurIPS}, 2021.

\bibitem{yuan2021hrformer}
Yuhui Yuan, Rao Fu, Lang Huang, Weihong Lin, Chao Zhang, Xilin Chen, and
  Jingdong Wang.
\newblock Hrformer: High-resolution transformer for dense prediction.
\newblock In {\em NeurIPS}, 2021.

\bibitem{yun2019cutmix}
Sangdoo Yun, Dongyoon Han, Seong~Joon Oh, Sanghyuk Chun, Junsuk Choe, and
  Youngjoon Yoo.
\newblock Cutmix: Regularization strategy to train strong classifiers with
  localizable features.
\newblock In {\em ICCV}, 2019.

\bibitem{zhang2021vil}
Pengchuan Zhang, Xiyang Dai, Jianwei Yang, Bin Xiao, Lu Yuan, Lei Zhang, and
  Jianfeng Gao.
\newblock Multi-scale vision longformer: A new vision transformer for
  high-resolution image encoding.
\newblock In {\em ICCV}, 2021.

\bibitem{zhong2020random}
Zhun Zhong, Liang Zheng, Guoliang Kang, Shaozi Li, and Yi Yang.
\newblock Random erasing data augmentation.
\newblock In {\em AAAI}, 2020.

\bibitem{zhou2017ade20k}
Bolei Zhou, Hang Zhao, Xavier Puig, Sanja Fidler, Adela Barriuso, and Antonio
  Torralba.
\newblock Scene parsing through ade20k dataset.
\newblock In {\em CVPR}, 2017.

\bibitem{zhu2020deform-detr}
Xizhou Zhu, Weijie Su, Lewei Lu, Bin Li, Xiaogang Wang, and Jifeng Dai.
\newblock Deformable detr: Deformable transformers for end-to-end object
  detection.
\newblock In {\em ICLR}, 2021.

\end{thebibliography}
}

\end{document}